\documentclass[11pt]{article}

\usepackage[preprint]{acl}

\usepackage{times}
\usepackage{latexsym}

\usepackage[T1]{fontenc}

\usepackage[utf8]{inputenc}

\usepackage{microtype}

\usepackage{inconsolata}

\usepackage{graphicx}
\usepackage{booktabs}
\usepackage{multicol}
\usepackage{multirow}
\usepackage{array}
\usepackage{makecell}
\usepackage{booktabs, makecell, siunitx, threeparttable, caption}
\usepackage{amsmath}
\usepackage{mdframed}
\usepackage[most]{tcolorbox}
\usepackage{tikz}
\usetikzlibrary{shadows}
\usepackage{enumitem}
\usepackage{subcaption}
\usepackage[table]{xcolor}
\usepackage{algorithm}
\usepackage{algpseudocode}
\usepackage{amssymb}
\tcbuselibrary{skins, breakable}
\usepackage{array}
\usepackage{placeins}
\usepackage{supertabular}
\usepackage{nicematrix}
\usepackage{xcolor}

\title{LifeSim: Long-Horizon User Life Simulator for Personalized Assistant Evaluation}

\author{
 \textbf{Feiyu Duan\textsuperscript{1}},
 \textbf{Xuanjing Huang\textsuperscript{3}},
 \textbf{Zhongyu Wei\textsuperscript{1,2,4,5\thanks{Corresponding author.}}},
\\
 \textsuperscript{1}School of Data Science, Fudan University \\
 \textsuperscript{2}Shanghai Innovation Institute \\
 \textsuperscript{3}College of Computer Science and Artificial Intelligence, Fudan University \\
 \textsuperscript{4}MOE laboratory for National Development and Intelligent Governance, Fudan University \\
 \textsuperscript{5}Research Institute of Intelligent Complex Systems, Fudan University \\
  \texttt{fyduan25@m.fudan.edu.cn}\quad 
  \texttt{\{xjhuang,zywei\}@fudan.edu.cn}
}

\begin{document}
\maketitle

\begin{abstract}

The rapid advancement of large language models (LLMs) has accelerated progress toward universal AI assistants. However, existing benchmarks for personalized assistants remain misaligned with real-world user-assistant interactions, failing to capture the complexity of external contexts and users' cognitive states. To bridge this gap, we propose \textbf{LifeSim}, a user simulator that models user cognition through the Belief-Desire-Intention (BDI) model within physical environments for coherent life trajectories generation, and simulates intention-driven user interactive behaviors.
Based on LifeSim, we introduce \textbf{LifeSim-Eval}, a comprehensive benchmark for multi-scenario, long-horizon personalized assistance. LifeSim-Eval covers 8 life domains and 1,200 diverse scenarios, and adopts a multi-turn interactive method to assess models' abilities to complete explicit and implicit intentions, recover user profiles, and produce high-quality responses. Under both single-scenario and long-horizon settings, our experiments reveal that current LLMs face significant limitations in handling implicit intention and long-term user preference modeling.~\footnote{Our code and data are available at \href{https://github.com/dfy37/lifesim}{https://github.com/\\dfy37/lifesim}.}

\end{abstract}

\section{Introduction}

\begin{figure}[t]
    \centering
    \includegraphics[width=1.0\linewidth]{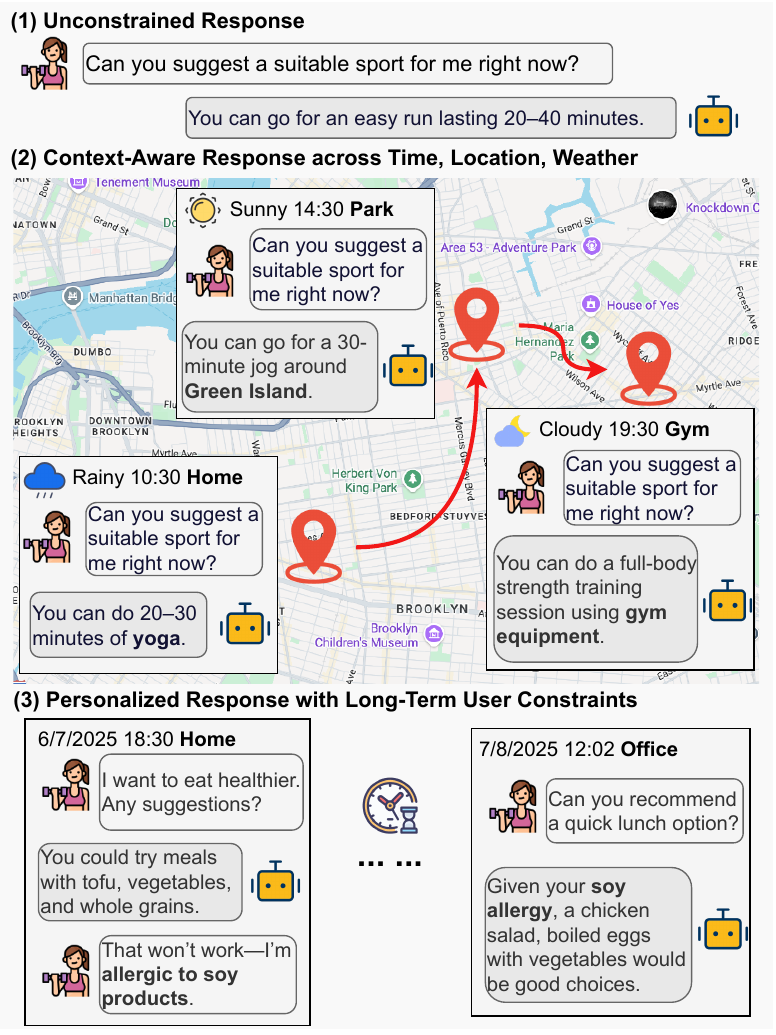}
    \caption{Illustration of personal AI assistance grounded in long-horizon spatiotemporal context. User behaviors evolve with external environment, while reflecting stable personal traits. Effective response requires models to adjust their strategies to current context while leveraging interaction history to infer personal states.}
    \label{fig:illustration}
\end{figure}

The rapid advancement of large language models (LLMs) has substantially expanded the capabilities of AI assistants across diverse scenarios and tasks, making the vision of a "Jarvis-like" universal digital assistant increasingly achievable~\cite{raza2025industrial}. Recent research has proposed optimizations from multiple perspectives~\cite{yuan2025personalized, jiang2025know, zhao2025teaching} to improve not only the model's ability to address complex and knowledge-intensive tasks, but also its social intelligence.~\cite{mathur2024advancing}. However, a clear gap persists between current evaluation frameworks and real-world scenarios, constraining advances in personalized intelligence~\cite{jiang2025know, zhao2025llms, kim2024dialsim}.



\begin{figure*}[t]
    \centering
    \includegraphics[width=1.0\linewidth]{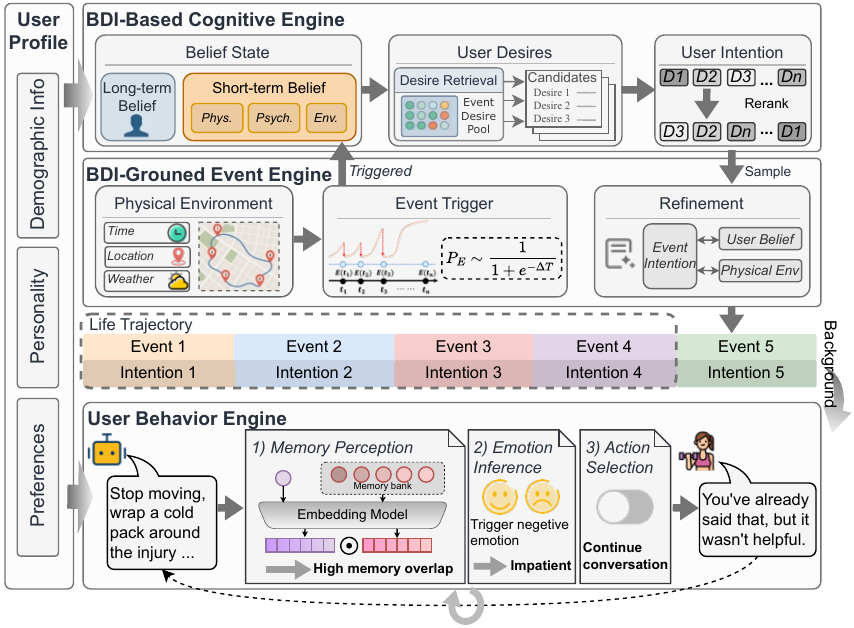}
    \caption{Overview of the LifeSim framework. For each target user, the user profile consists of demographic attributes, personality traits, and long-term preferences, which together contribute to the long-term belief state. The BDI-based cognitive engine and the event engine jointly generate user intentions by integrating subjective belief states with physical environments. The user behavior engine then produces conversations by modeling memory perception, emotion inference, and action selection.}
    \label{fig:framework}
\end{figure*}

As illustrated in Figure~\ref{fig:illustration}, ideal realistic user-assistant interactions fundamentally differ from isolated question answering and involve two critical dimensions of complexity. 
(1) \textbf{Complex external environment}: user needs vary in terms of situational factors, such as time, location, weather, and ongoing life events~\cite{kim2025cupid, fan2025ai};
(2) \textbf{Dynamic cognitive states of users}: user intentions arise from internal cognitive states, jointly shaped by evolving life experience, as well as relatively stable personalities and preferences~\cite{wu2024longmemeval}.
Real-world user data are constrained by privacy and ethical considerations, and publicly available interaction logs spanning multiple years and diverse scenarios remain extremely scarce~\cite{xu2022globem}. Consequently, existing benchmarks are often forced to rely on static or short-context datasets, which makes it difficult to faithfully reflect the dynamic nature of how users and assistants interact in real-world settings. Therefore, establishing a realistic testbed with long-term user-assistant interactions at scale poses a fundamental problem. 

To bridge the gap, we introduce the \textbf{Long-horizon user \underline{Life} \underline{Sim}ulator (LifeSim)}, a high-fidelity framework that supports the behavior simulation of diverse users across life trajectories constrained on both internal cognitive model and external environment. For user cognitive modeling, we adopt the Belief-Desire-Intention (BDI) model construct the user's internal reasoning process.
To simulate the complicated external environment, LifeSim integrates an event engine to generate life trajectories under the guidance of the BDI model. To realize user interaction behavior, LifeSim incorporates a user behavior engine, which generates coherent responses for the target user aligned with its internal cognition and external contexts. To ensure population diversity, we construct a million-level user pool, with rich persona attributes about demographics, personality traits and preferences.

Building upon LifeSim, we develop \textbf{LifeSim-Eval}, a new benchmark for the evaluation of a model's capabilities in long-term individualized interactions, comprising 1,200 scenarios across 8 common life domains. LifeSim-Eval employs an online interaction protocol to evaluate a model's ability to recognize user's intentions, to provide preference-aligned responses, and to recall user preferences throughout long-term interactions. We conduct systematic experiments on more than 10 open-source and closed-source models. The results show that: (i) current LLMs demonstrate strong performance in handling explicit intentions, yet struggle with implicit intentions and long-horizon user modeling;
(ii) simple profile memory offers limited benefits, indicating that effective personalization requires stable preference reasoning beyond simple retention.




\section{LifeSim Framework}
Figure~\ref{fig:framework} illustrates the overall architecture of \textbf{LifeSim}, which including four main components: \textit{BDI-Based Congnitive Engine}, \textit{BDI-Grounded Event Engine}, \textit{User Behavior Engine} and \textit{User Profile Pool}. 
For each simulation, we first sample a user profile from profile pool to initialize the long term belief state. Given a corresponding spatial trajectory for the target user, LifeSim constructs a life trajectory, where each possible service point is associated with a \textit{user intention} and a corresponding \textit{life event}. Algorithm~\ref{alg:event_gen} summarizes the life trajectory generation process. User behavior engine simulates user responses based on the generated background. We will detail the architecture and the generation process in the following sub-sections. 



\subsection{BDI-based Cognitive Engine}
\label{sec:BDI-based Cognitive Engine}
To model the internal reasoning process of a target user, we adopt the Belief-Desire-Intention (BDI) model~\cite{bratman1987intention}, a psychological model of human practical reasoning. 

\paragraph{Belief State} 

Belief summarizes the information relevant to the user's current decision state, including \textit{long-term belief}, which corresponds to the user profile; and \textit{short-term belief}, which reflects the user’s cognition of the current situation. Given a physical environment $Env$, we use the LLM to generate event hypotheses $H$ based on the user’s long-term belief and recent life experiences, which serve as the user’s short-term belief. $H$ characterize the events that are reasonable and likely to occur in the current situation. Event hypothesis is designed following Lewin's equation in social psychology~\cite{heidbreder1937lewin}, $B = f(P, Env)$, which highlights that user behavior $B$ emerges from the joint influence of person $P$ and environment $Env$.


\paragraph{User Desires}
Desires are conceptualized as a set of intention candidates conditioned on the user's beliefs. In practice, we construct a \emph{Desire Pool} that includes a wide range of potential user intentions. At each service point, we retrieve 9 candidate intentions from the pool conditioned on user's belief states. 
For desire pool construction, each intention is associated with a corresponding latent event context, which characterizes the real-world situation in which this intention is most likely to arise. To achieve a broad, yet practical coverage, we organize the collected intentions according to a set of high-level life domains adapted from the World Health Organization Quality of Life (WHOQOL) framework~\cite{whoqol1995world}. Construction details and statistics are provided in Appendix~\ref{subsec:event_intent_pool_cons}



\paragraph{User Intention}
User intention denotes a future-directed commitment selected from desires, indicating what the user is prepared to pursue next. Retrieved user desires are first reranked conditioned on the user's belief state and the current environment. During reranking, candidates that are logically inconsistent with the user's belief and physical environment are filtered out. The remaining candidates are then transformed into a probability distribution using a softmin function over their ranks $r$:
\begin{equation}
    P(I_i) = \frac{\exp(-r_i)}{\sum_k \exp(-r_k)}
\end{equation}
from which the final intention is sampled.

\subsection{BDI-Grounded Event Engine}
\label{sec:BDI-Grounded Event Engine}
The event engine constructs users' life trajectories grounded in realistic physical environments.

\paragraph{Physical Environment}
We represent the environment as $Env(T, L, W)$, where $T$ denotes time, $L$ denotes location, and $W$ denotes weather. To ensure realism, the environment is grounded in user mobility trajectories, allowing intentions to emerge in a manner consistent with plausible daily routines. We collect 3,374 user trajectories that span 251 Points Of Interest (POI). Detailed procedures for mobility data collection and profile matching are provided in Appendix~\ref{subsec:user_travel_traj}.

\paragraph{Event Trigger}
Since users might not ask for a service at every trajectory point, we introduce an event trigger to determine whether an event would be generated. 
The occurrence probability of an event at the timestamp $t$, denoted as $P_E$, is defined as a function of the elapsed time $\Delta T$ since the last triggered event, i.e., $P_E = f(\Delta T)$, where $f(z) = \frac{1}{1+e^{- z}}$ is specified as a logistic function. We then perform Bernoulli sampling based on $P_E$ to determine whether an event is triggered at the current point. If an event is triggered, the environmental information $Env$ is passed to the cognitive engine to construct the user’s short-term belief.

\paragraph{Refinement}
After an intention is generated in the cognitive engine, the event engine leverages its associated event context and grounds it within the current environmental setting. As the retrieved event and intention may not fully align with the specific environmental conditions or user attributes, the event engine further refines the details to ensure better consistency.

\subsection{User Behavior Engine}
\label{sec:User Behavior Engine}



Based on the given intention, the user behavior engine generates user conversational behaviors. During simulation, each user turn is modeled through four sequential stages:
\textbf{(1) Memory Perception.} Whenever an assistant response $r$ contains an informational value, it is abstracted into a $\langle \text{query}, \text{value} \rangle$ pair and stored as a memory entry. Simultaneously, the response is compared with existing memories using semantic similarity; if the similarity exceeds a predefined threshold $\theta$\footnote{We take $\theta=0.7$ during experiments.}, the system triggers a negative memory perception.
\textbf{(2) Emotion Inference.} Following the emotion-chain reasoning paradigm introduced in AnnaAgent~\cite{wang2025annaagent}, we adopt the emotion taxonomy from GoEmotions~\cite{kalyuga2009expertise} and predict the user's emotional state prior to generating each response.
\textbf{(3) Action Selection.} Based on the inferred information, the agent determines whether the user continues the interaction.
\textbf{(4) Response Generation.} The final user response is then generated conditioned on memory perception, emotion inference, and the selected action.

\begin{table}[t]
    \centering
    \scalebox{1.0}{
    \begin{tabular}{l c}
        \toprule
            \textbf{Intention Type} & \textbf{Percentage(\%)} \\
        \midrule
            Problem-Solving            & 43.4 \\
            Informational              & 17.1 \\
            Educational                & 11.5 \\
            Personal Interaction       & 10.9 \\
            Technical and Professional & 10.2 \\
            Creative                   & 6.8  \\
        \bottomrule
    \end{tabular}
    }
    \caption{Distribution of intention types in LifeSim-Eval.}
    \label{tab:intent_distribution}
\end{table}

\subsection{User Profile}
\label{sec:user_profile}
Each user is modeled along three dimensions: \textit{\textbf{(1) Demographic attributes}} capture coarse-grained background information, such as age, gender, and education level. \textit{\textbf{(2) Personality traits}} are characterized using the Big Five theory, represented along five canonical dimensions. \textit{\textbf{(3) Long-term preferences}} reflect persistent inclinations toward specific activities, needs, or conversational styles, and remain stable across multi-scenarios.

To construct a large-scale user pool, we leverage the SocioVerse dataset~\cite{zhang2025socioverse}, which is sourced from Twitter. The pool is refined by filtering out suspected bot accounts, and each remaining user is annotated with 15 core demographic attributes. Meanwhile, we adopt the AlignX dataset~\cite{li20251}, which draws on Maslow's hierarchy of needs and Murray's system of needs to define a 90-dimensional personality and preference space spanning psychological and behavioral aspects. We retain the personality traits and 40 additional dimensions most relevant to everyday life routines and social interactions. Consequently, we construct a user pool of 1M individuals.

\section{LifeSim-Eval Benchmark}



\begin{table*}[t]
    \centering
    \renewcommand{\arraystretch}{1.0}
    \scalebox{0.75}{
    \begin{NiceTabular}{lcccccccc}
        \toprule
            \multirow{2}{*}{\textbf{Model}} 
            & \multicolumn{2}{c}{\textbf{Intent Recognition}}
            & \multicolumn{2}{c}{\textbf{Intent Completion}}
            & \multirow{2}{*}{\textbf{Naturalness}} 
            & \multirow{2}{*}{\textbf{Coherence}} 
            & \multirow{2}{*}{\textbf{Profile Recov.}} 
            & \multirow{2}{*}{\textbf{Persona Align.}} \\
        \cmidrule(lr){2-3} \cmidrule(lr){4-5}
            & \textit{\textbf{explicit}} & \textit{\textbf{implicit}} & \textit{\textbf{explicit}} & \textit{\textbf{implicit}} 
            & & & & \\
        \midrule
            \rowcolor{gray!15}
            \Block[c]{1-9}{\textbf{\textit{Close-source}}} \\
        \midrule
            GPT-5 & 79.5 & 52.2 & \textbf{76.9} & 48.9 & 83.9 & \textbf{97.5} & 60.3 & 74.6 \\
            GPT-4o            & 79.0 & 52.6 & 72.1 & 48.1 & 81.4 & 90.0 & 60.5 & 74.1 \\
            Claude-Sonnet-4.5 & 76.1 & 49.2 & 73.0 & 46.5 & 89.8 & 96.9 & 60.9 & \textbf{75.5} \\
        \midrule
            \rowcolor{gray!15}
            \Block[c]{1-9}{\textbf{\textit{Open-source}}} \\
        \midrule
            DeepSeek-V3.2 & 78.6 & 54.6 & 73.5 & 50.8 & \textbf{85.2} & 91.5 & \textbf{65.0} & \textbf{75.5}   \\
            DeepSeek-V3.2 Thinking & \textbf{80.6} & \textbf{59.3} & 75.8 & \textbf{58.2} & 84.1 & 92.1 & 63.4 & 74.0 \\
            Llama3.1-it 8B      & 70.3 & 40.3 & 58.4 & 33.8 & 69.3 & 72.8 & 58.5 & 73.8 \\
            Llama3.1-it 70B    & 71.7 & 41.4 & 60.9 & 35.5 & 79.6 & 84.8 & 60.4 & 73.5 \\
            Gemma3-it 12B       & 67.8 & 35.8 & 57.0 & 30.0 & 67.3 & 71.0 & 60.7 & 73.5 \\
            Gemma3-it 27B       & 72.7 & 44.2 & 65.4 & 39.3 & 78.4 & 81.3 & \textbf{61.8} & 75.6 \\
            gpt-oss 20B         & 76.9 & 43.7 & 72.6 & 41.1 & 83.4 & 91.0 & 57.1 & 75.2 \\
            gpt-oss 120B        & 76.4 & 44.3 & \textbf{73.3} & 42.6 & \textbf{85.8} & \textbf{93.0} & 59.8 & 74.2 \\
            Qwen3 8B         & 75.7 & 44.4 & 64.4 & 38.4 & 75.0 & 83.3 & 60.5 & 73.5 \\
            Qwen3 14B        & \textbf{77.0} & 46.3 & 68.3 & 41.3 & 79.1 & 86.8 & \textbf{61.8} & 74.7 \\
            Qwen3 32B        & 76.6 & \textbf{51.2} & 70.3 & \textbf{47.4} & 83.9 & 89.9 & 61.7 & \textbf{75.8} \\
        \bottomrule
    \end{NiceTabular}
    }
    \caption{Evaluation results of open-source and proprietary LLMs on LifeSim-Eval, measuring intent recognition and completion, response quality, and user modeling with personalized responses. All values are linearly applied to the [0,100] scale. Bold values indicate the highest scores among open-source or proprietary models. Abbreviations: Profile Recov. = Profile Recovery, Persona Align. = Persona Alignment.}
    \label{tab:main_results}
\end{table*}

Based on LifeSim, we introduce LifeSim-Eval, a benchmark for long-horizon personalized assistants that evaluates AI performance in dynamically evolving interaction scenarios. Such scenarios inherently require assistants to handle user intentions. However, intention inference is non-trivial, as some intentions are directly expressed in the current utterance, while others remain implicit~\cite{allott2013relevance}. These implicit constraints may originate from local scenario context or from user preferences and states accumulated over time. LifeSim-Eval therefore evaluates whether assistants can satisfy both explicit and implicit intentions. More details are provided in Appendix~\ref{subsec:LifeSim-Eval_Details}.

\paragraph{Single Scenario Setting}
LifeSim-Eval first evaluates assistant behavior in each single scenario without access to interaction history, where implicit intentions are mainly induced by user profile or environmental context. We assess whether the assistant correctly recognizes and fulfills both explicit and implicit intentions using \textit{Intent Recognition} and \textit{Intent Completion}. To evaluate assistant response qualities and user modeling abilities, we further report four auxiliary metrics: \textit{Naturalness}, \textit{Coherence}, \textit{Preference Recovery}, and \textit{Persona Alignment}. Metric definitions are provided in Appendix~\ref{subsec:LifeSim-Eval_Details}. We adopt LLM-as-Judge protocol and report the average scores from GPT-4o\footnote{We use the version in 2025/11/20}, Qwen3-32B~\cite{yang2025qwen3}, and Llama3.1-it 70B~\cite{dubey2024llama}.

\paragraph{Long-Horizon Setting}
We further evaluate long-term performance by conditioning on LifeSim-simulated dialogue histories. 
User preferences are revealed in the dialogue of the first scenario, while the assistant is evaluated in the final scenario of the trajectory. While explicit intentions are specified in the current scenario, implicit intentions must be inferred from the user's belief state, which integrates stable preferences from the first scenario and dynamic states derived from historical life events. In this setting, we use \textit{Intent Completion} as the evaluation metric.

\paragraph{Construction of the Benchmark}
LifeSim-Eval includes 120 users, each associated with a sequence of 10 events, yielding 1,200 scenarios evenly distributed across eight life domains, with no repeated events per user. We adopt DeepSeek-V3.2 as the backbone model for cognitive engine and event engine, and generate the 1,200 scenarios through LifeSim. In the Single-Scenario setting, each scenario allows up to 20 turns, and user sampling is balanced using Iterative Proportional Fitting (IPF). In the Long-Horizon setting, interactions are limited to three turns. We simulate dialogue histories for 100 users using Qwen-32B as the user model and DeepSeek-Chat as the assistant, with event sequence lengths ranging from 1 to 10, and a total conversation history exceeding 14K tokens. All histories and evaluation scenarios are manually verified for temporal and logical consistency. Table~\ref{tab:intent_distribution} summarizes the distribution of intention types. More details are provided in Appendix~\ref{subsec:LifeSim-Eval_Details}.

\section{Main Experiments}
\label{sec:main_experiments}

Based on LifeSim-Eval, we assess the capability of mainstream LLMs to deliver personalized services. 

\paragraph{Experimental Setup} To reduce evaluation cost, all experiments employ the open-source Qwen3-32B model as the user agent. The evaluated assistant models span multiple series and sizes, including close-source models (GPT-5, GPT-4o, Claude Sonnet 4.5, DeepSeek-V3.2, DeepSeek-V3.2 Thinking), and open-source models (Qwen3-8B/14B/32B~\cite{yang2025qwen3}, Gemma3-it 12B/27B~\cite{team2025gemma}, Llama3.1-it 8B/70B~\cite{dubey2024llama}, and gpt-oss-20B/120B~\cite{agarwal2025gpt}). All experiments are conducted on 8 NVIDIA RTX 4090 GPUs. We use vLLM~\cite{kwon2023efficient} for model serving during inference. The sampling temperature is fixed at 1.0 throughout all experiments. 

\begin{figure}[t]
    \centering
    \includegraphics[width=\linewidth]{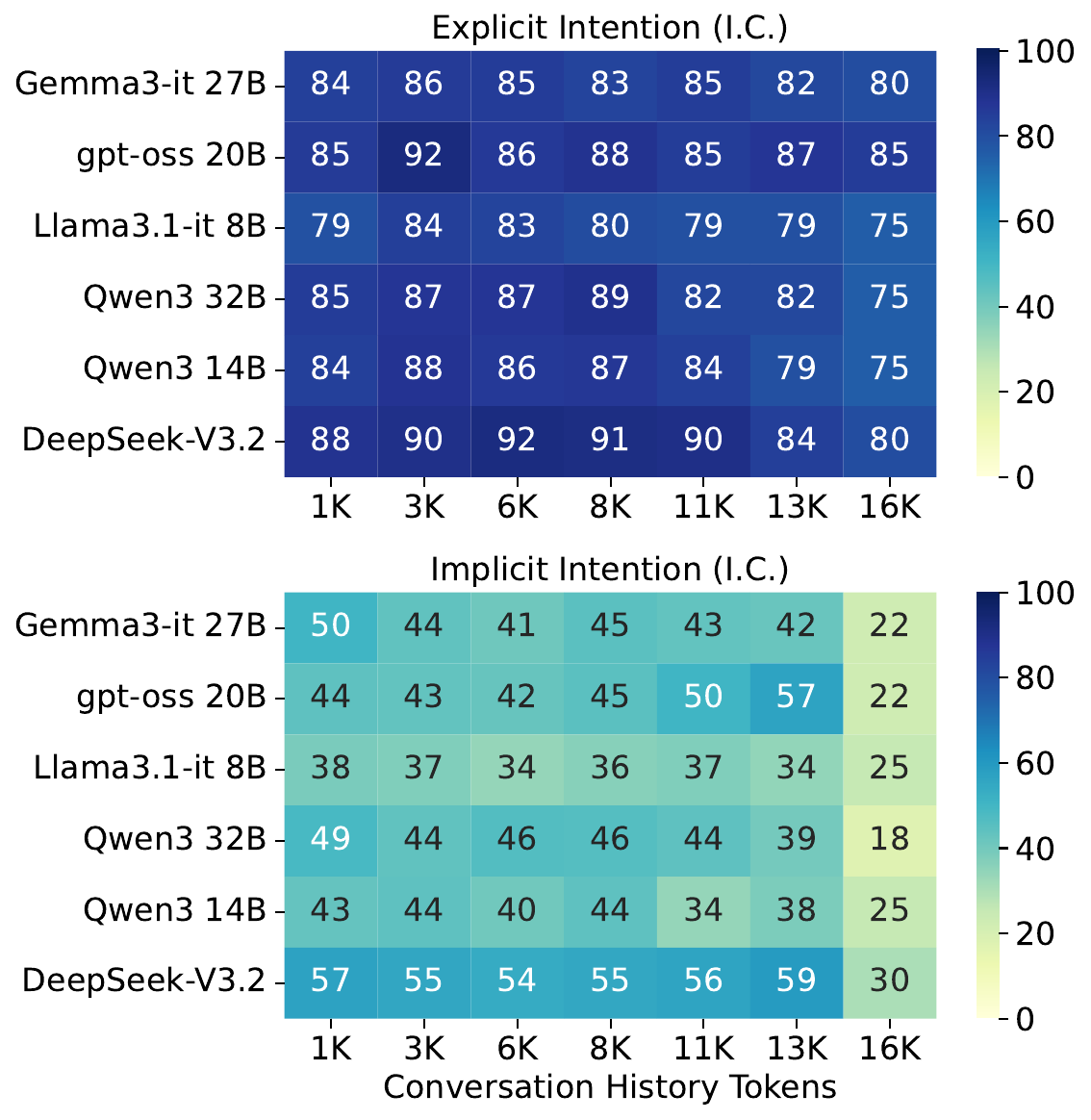}
    \caption{Long-horizon intention completion performance across different assistant models. The heatmaps report intention completion (I.C.) scores with respect to conversation length.}
    \label{fig:long-horizon setting}
\end{figure}

\begin{figure}[t]
    \centering
    \begin{subfigure}{0.49\linewidth}
        \centering
        \includegraphics[width=\linewidth]{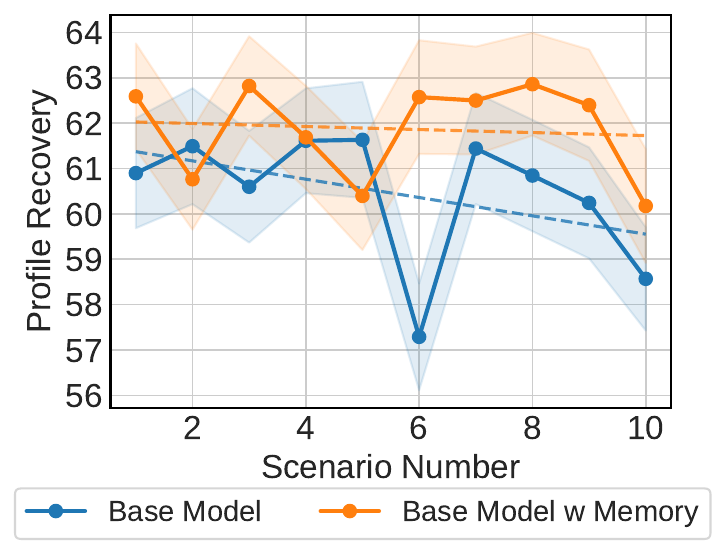}
        \caption{Qwen3 8B}
        \label{fig:Qwen3 8B}
    \end{subfigure}
    \hfill
    \begin{subfigure}{0.49\linewidth}
        \centering
        \includegraphics[width=\linewidth]{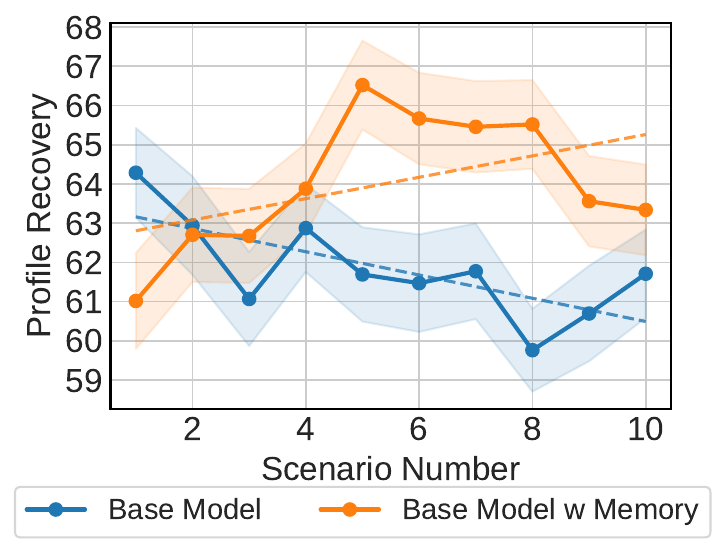}
        \caption{Qwen3 14B}
        \label{fig:Qwen3 14B}
    \end{subfigure}
    \\
    \begin{subfigure}{0.49\linewidth}
        \centering
        \includegraphics[width=\linewidth]{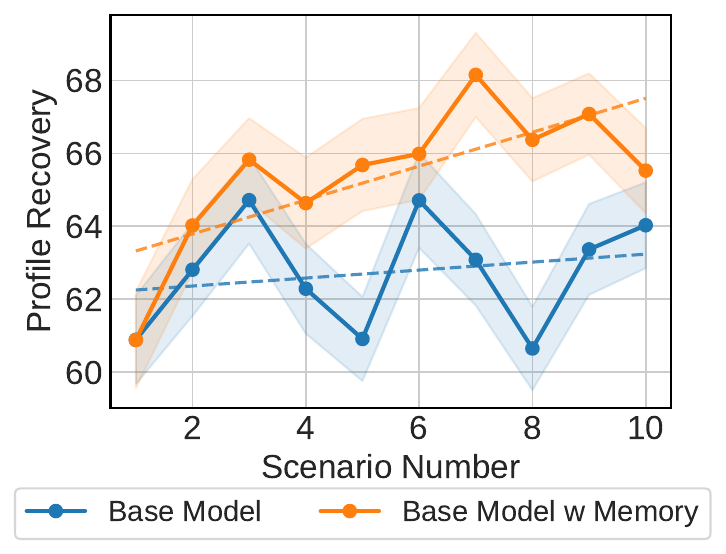}
        \caption{Gemma3-it 12B}
        \label{fig:Gemma3-it 12B}
    \end{subfigure}
    \hfill
    \begin{subfigure}{0.49\linewidth}
        \centering
        \includegraphics[width=\linewidth]{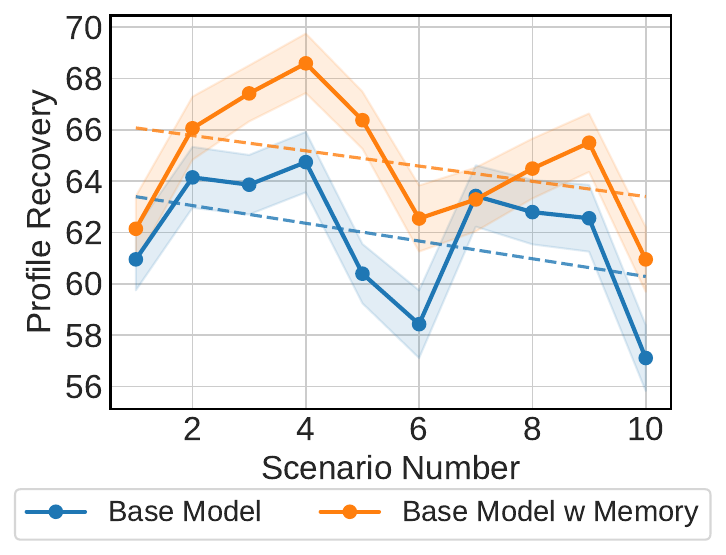}
        \caption{Gemma3-it 27B}
        \label{fig:Gemma3-it 27B}
    \end{subfigure}
    \caption{Performance of user preference recovery within life event sequences. The dashed line represents the regression curve fitted by linear regression.}
    \label{fig:preference_prediction_line}
\end{figure}

\paragraph{Performance in Single Scenario Setting} 

As shown in Table \ref{tab:main_results}, most models perform well on explicit intent recognition, while exhibiting a performance gap of more than 20 points on implicit intent recognition. This indicates that although current models can reliably address users' explicit queries, they remain limited in uncovering latent needs-an ability that is essential for realistic interactive experiences. 
In terms of overall dialogue quality, gpt-oss-120B demonstrates the best performance. However, DeepSeek-V3.2 achieves both a higher preference recovery score and the best persona alignment performance. Notably, although the Gemma family models achieve strong preference recovery accuracy, their persona alignment scores lag behind, suggesting limited ability to leverage persona traits to generate personalized responses.

Additionally, with the exception of DeepSeek models, proprietary models generally outperform open-source models, with notable advantages in implicit intention recognition and completion. Model scaling also leads to consistent performance gains across most of the metrics. Moreover, DeepSeek-V3.2 Thinking shows clear improvements over its base model in intention-related tasks, indicating that reasoning models can better capture both explicit user queries and latent user needs.


\paragraph{Performance in Long-Horizon Setting}

Results in Table~\ref{fig:long-horizon setting} show that, under long interaction histories, models retain strong performance on explicit intentions, with intention completion remaining stable as context length increases. In contrast, implicit intentions, which require integrating long-term dialog history and latent user preferences, exhibit substantially weaker performance. Moreover, implicit intention completion shows a clear decline as the conversation history grows longer. These results indicate that while current models can handle explicitly stated intention even in long contexts, they struggle to reason over long-horizon evidence to infer and satisfy implicit user intentions.

\paragraph{Profile Recovery in Long-Horizon Setting} We further investigate whether models can consistently model users' preferences over multi-scenario interactions. To this end, we introduce a simple profile memory mechanism: after each scenario, the model summarizes or refines the user's preferences, and this structured preference summary is then provided in subsequent scenarios. We evaluate each model's preference recovery accuracy across the whole trajectory for each user. 

Results in Figure \ref{fig:preference_prediction_line} show that incorporating preference memory leads to consistent benefits. However, Gemma3-it-12B and Qwen3-14B exhibit a clear upward trend when equipped with memory, whereas Qwen3-8B and Gemma3-it-27B remain nearly flat or even slightly decline. These discrepancies suggest that a simple external memory mechanism alone is insufficient to guarantee robust long-term preference modeling. Achieving reliable and incremental long-horizon user modeling performance requires not only the ability to retain preference information but also the capability to perform stable preference reasoning throughout prolonged interactions.

\section{Further Analysis}
\label{further_analysis}

\paragraph{Impact of Different Themes and Types}

\begin{figure}[t]
    \centering
    \begin{subfigure}{0.49\linewidth}
        \centering
        \includegraphics[width=\linewidth]{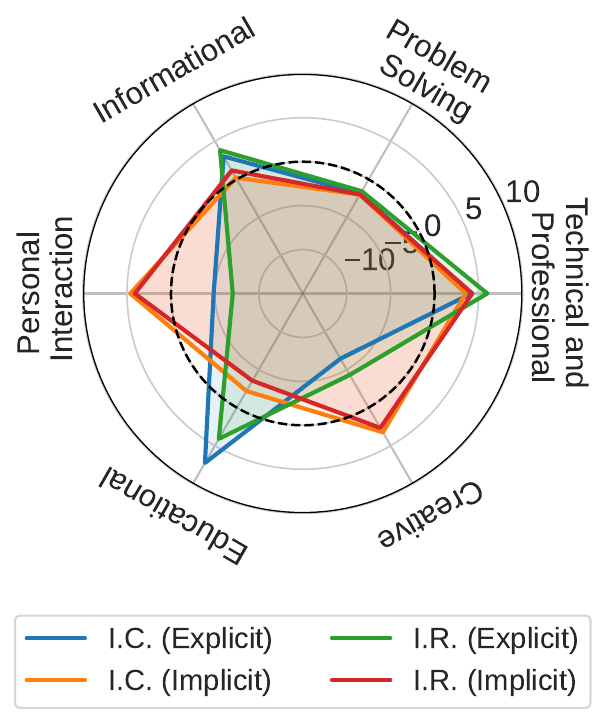}
        \caption{Qwen3 8B}
        \label{fig:Qwen3 8B}
    \end{subfigure}
    \hfill
    \begin{subfigure}{0.49\linewidth}
        \centering
        \includegraphics[width=\linewidth]{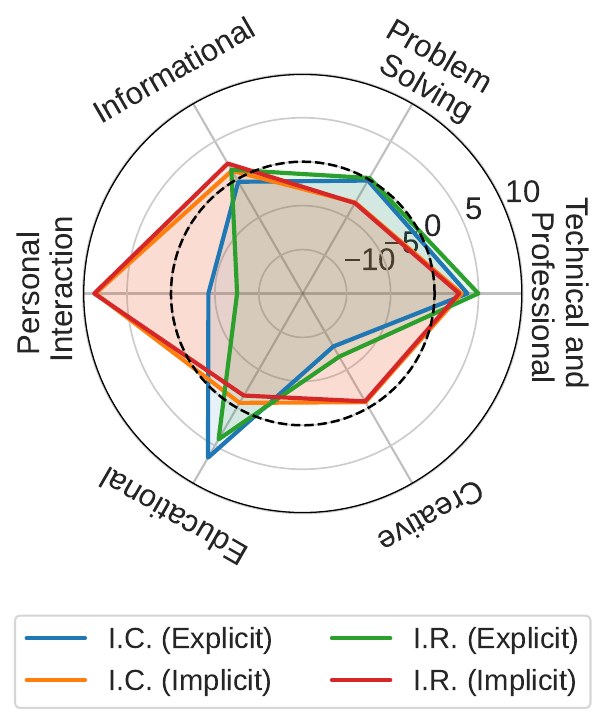}
        \caption{Qwen3 14B}
        \label{fig:Qwen3 14B}
    \end{subfigure}
    \\
    \begin{subfigure}{0.49\linewidth}
        \centering
        \includegraphics[width=\linewidth]{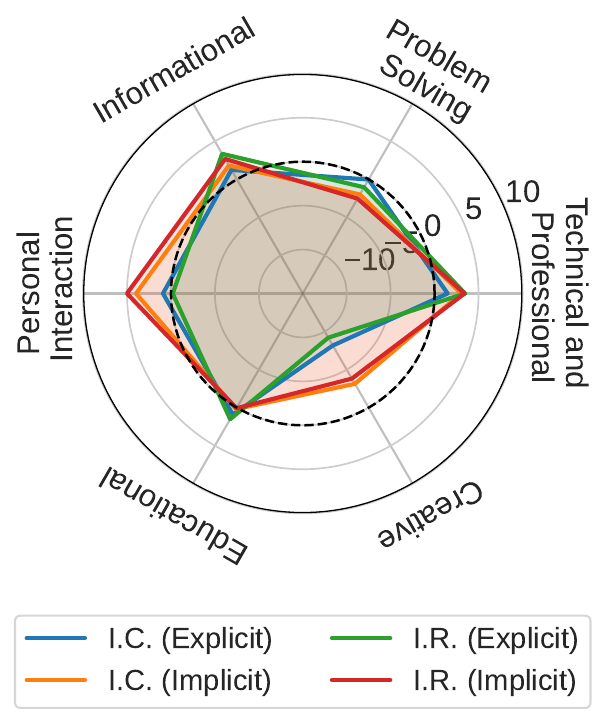}
        \caption{Gemma3-it 12B}
        \label{fig:Gemma3-it 12B}
    \end{subfigure}
    \hfill
    \begin{subfigure}{0.49\linewidth}
        \centering
        \includegraphics[width=\linewidth]{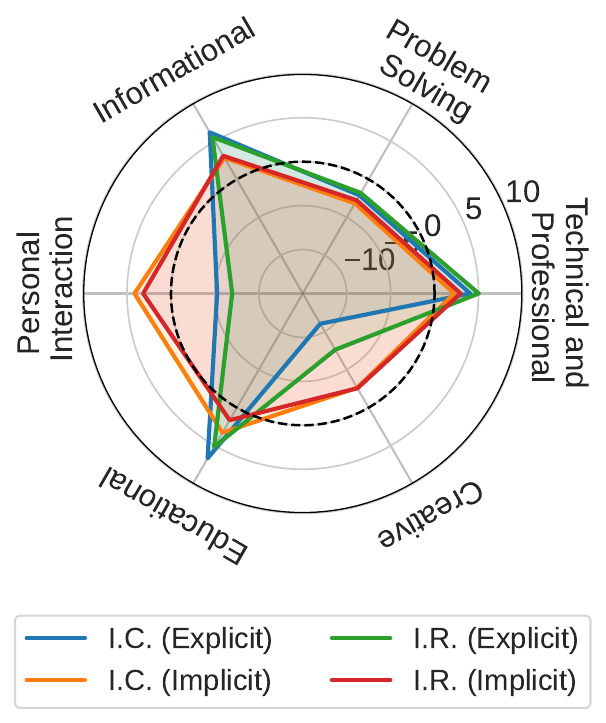}
        \caption{Gemma3-it 27B}
        \label{fig:Gemma3-it 27B}
    \end{subfigure}
    \caption{Relative model performance across different intention types. Abbreviations: I.R. = Intent Recognition, I.C. = Intent Completion}
    \label{fig:intent_types}
\end{figure}

\begin{figure}[t]
    \centering
    \begin{subfigure}{0.49\linewidth}
        \centering
        \includegraphics[width=\linewidth]{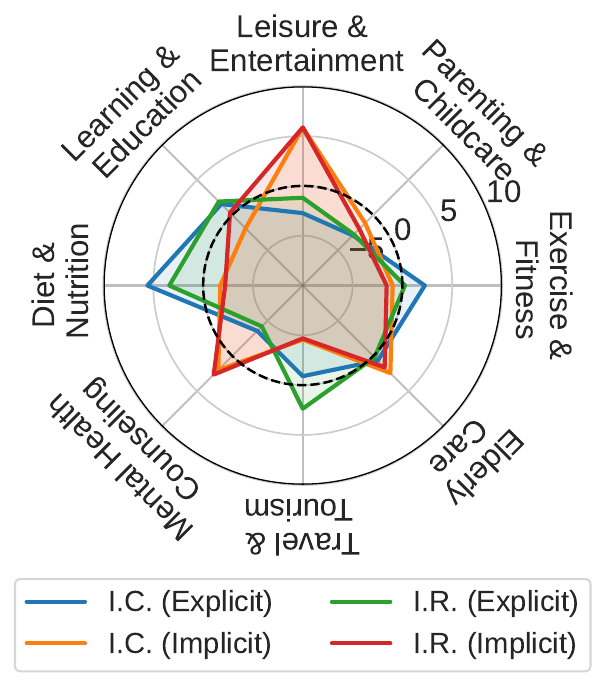}
        \caption{Qwen3 8B}
        \label{fig:Qwen3 8B}
    \end{subfigure}
    \hfill
    \begin{subfigure}{0.49\linewidth}
        \centering
        \includegraphics[width=\linewidth]{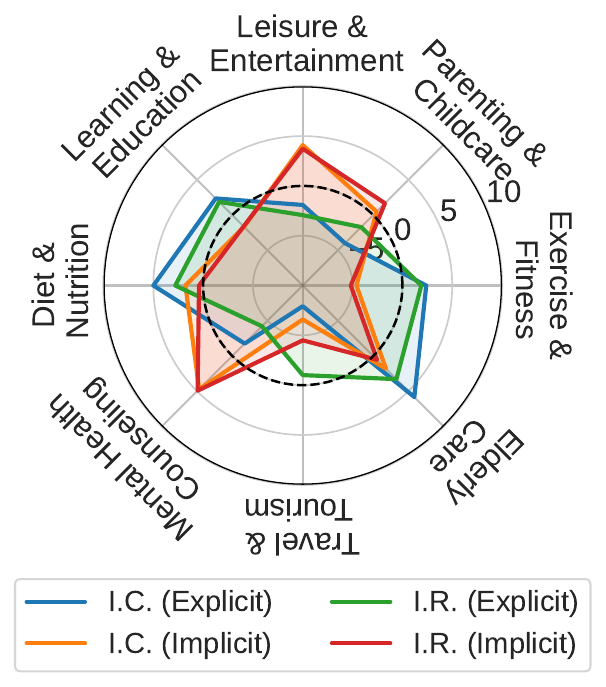}
        \caption{Qwen3 14B}
        \label{fig:Qwen3 14B}
    \end{subfigure}
    \\
    \begin{subfigure}{0.49\linewidth}
        \centering
        \includegraphics[width=\linewidth]{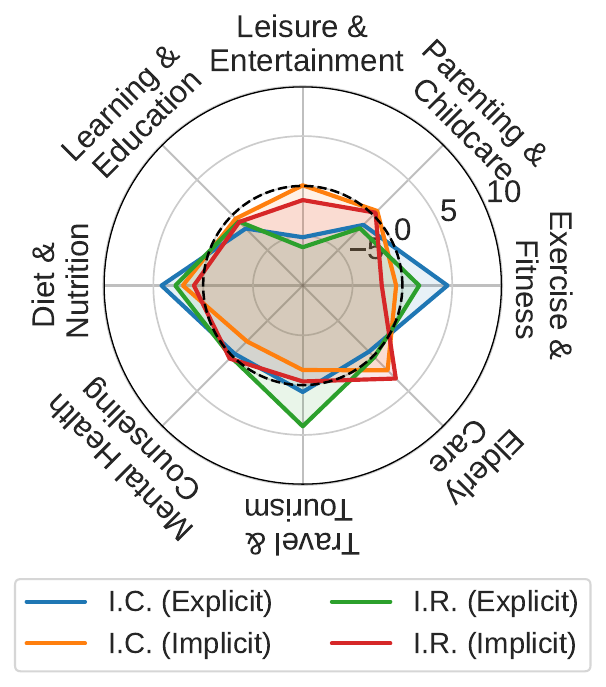}
        \caption{Gemma3-it 12B}
        \label{fig:Gemma3-it 12B}
    \end{subfigure}
    \hfill
    \begin{subfigure}{0.49\linewidth}
        \centering
        \includegraphics[width=\linewidth]{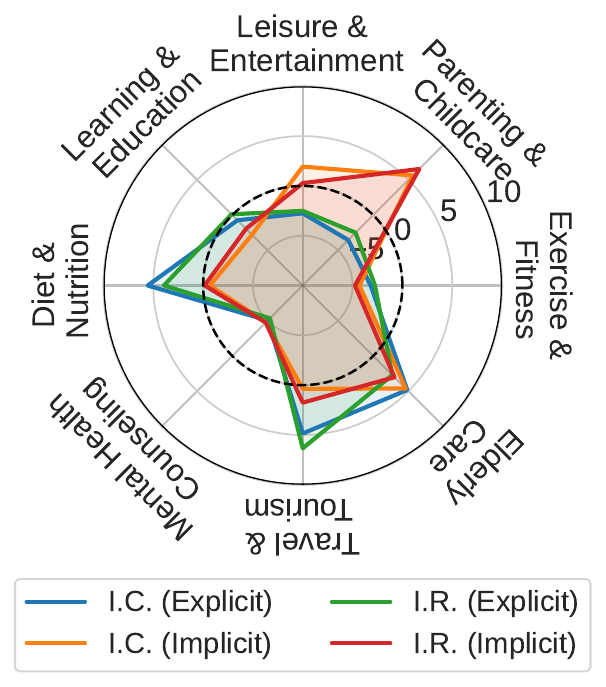}
        \caption{Gemma3-it 27B}
        \label{fig:Gemma3-it 27B}
    \end{subfigure}
    \caption{Relative model performance across different intention themes. Abbreviations: I.R. = Intent Recognition, I.C. = Intent Completion}
    \label{fig:intent_themes}
\end{figure}

\begin{table}[t]
    \centering
    \renewcommand{\arraystretch}{1.2}
    \scalebox{0.76}{
    \begin{NiceTabular}{lcccc}
        \toprule
            \textbf{Model} 
            & \textbf{I.A.} & \textbf{P.C.} & \textbf{C.R.} & \textbf{N.} \\
        \midrule
            \rowcolor{gray!15}
            \Block[c]{1-5}{\textbf{\textit{LLm-as-Judge}}} \\
        \midrule
            GPT-4o        & 97.9 & 94.0 & 95.6 & 99.5 \\
            DeepSeek-V3.2 & \textbf{98.4} & \textbf{97.1} & \textbf{96.8} & 99.9 \\
            Qwen3 32B     & 96.7 & 96.0 & 93.6 & \textbf{100.0} \\
        \midrule
            Qwen3 32B \textit{\small w/o Memory Perception}
                           & 96.5 & 86.0 & 91.0 & 99.9 \\
            Qwen3 32B \textit{\small w/o Emotion Inference}
                           & 95.6 & 87.1 & 90.3 & 99.9 \\
        \midrule
            \rowcolor{gray!15}
            \Block[c]{1-5}{\textbf{\textit{Human Evaluation}}} \\
        \midrule
            GPT-4o        & 94.7 & 93.6 & 94.7 & 91.1 \\
            DeepSeek-V3.2 & 95.8 & 92.7 & \textbf{96.2} & 91.3 \\
            Qwen3 32B     & \textbf{97.4} & \textbf{95.3} & \textbf{96.2} & \textbf{94.0} \\
        \midrule
            Qwen3 32B \textit{\small w/o Memory Perception}
                           & 97.1 & 94.4 & 95.3 & 89.6 \\
            Qwen3 32B \textit{\small w/o Emotion Inference}
                           & 96.7 & 93.8 & \textbf{96.2} & 92.2 \\
        \bottomrule
    \end{NiceTabular}
    }
    \caption{User Behavior Engine performance with difference model base across four dimensions. Bold values indicate the highest scores. Abbreviations: I.A.=Intent Alignment, P.C. = Persona Consistency, C.R. = Context Relevance, N. = Naturalness.}
    \label{tab:user_agent_performance}
\end{table}


We further analyze model performance across event themes and intention types. For each theme/type, we subtract the overall mean score to quantify its relative performance. Figure~\ref{fig:intent_types} and~\ref{fig:intent_themes} and reveal systematic variations across categories, indicating that model effectiveness is not uniform but depends on the nature of the underlying intention and interaction context. In particular, performance differs between settings dominated by explicit, task-driven needs and those requiring implicit, affective inference. This heterogeneity suggests that current models exhibit uneven robustness across service domains, motivating more fine-grained improvements in personalized assistant design.




\paragraph{User Behavior Engine Evaluation}

To evaluate the reliability of the LifeSim User Behavior Engine, we construct 300 scenarios using the LifeSim framework, each comprising a defined user profile, event background, and individual intention. The assistant model is fixed to DeepSeek-V3.2~\cite{deepseekai2024deepseekv32}, whereas the user model is instantiated with Qwen3-32B~\cite{yang2025qwen3}, DeepSeek-V3.2, and GPT-4o; ablation studies are additionally performed on Qwen3-32B. User simulation quality is assessed across four dimensions: (1) Intent Alignment, measuring whether generated dialogue accurately reflects the predefined intention; (2) Profile Consistency, examining whether utterances embody the specified persona; (3) Contextual Consistency, evaluating adherence to temporal, spatial, and situational constraints; and (4) Naturalness, assessing human-like fluency. Evaluation combines an LLM-as-Judge automatic protocol with human assessment on 30 sampled scenarios, rated independently by three proficient English annotators. See Appendix~\ref{sec:User Behavior Engine Evaluation} for more details.

As shown in Table \ref{tab:user_agent_performance}, all models achieve scores above 90 on most metrics, showing robust performance of our behavior engine. Ablation results further show performance drops across all dimensions when memory or emotion modules are removed, underscoring their importance in maintaining realism and long-range behavioral coherence.

\paragraph{Human-LLM Judge Alignment}
To quantify the alignment between LLM-based evaluation and human judgment, we randomly sample 150 simulation results from the single scenario setting, and compute Krippendorff's $\alpha$ between the LLM's scores (average from GPT-4o, Qwen3-32B, and Llama3.1-70B-Instruct) and those provided by three independent annotators across all the metrics. As shown in Table \ref{tab:Krippendorff}, the agreement coefficients are superior to 0.77 for all the annotators, with an average $\alpha$ of 0.80. These results indicate that the LLM achieves scoring consistency that is highly comparable to human annotators, showing the effectiveness of LifeSim-Eval.

\begin{table}[t]
    \centering
    \begin{tabular}{lc}
        \toprule
            \textbf{Pair} & \textbf{Krippendorff's $\alpha$} \\
        \midrule
            LLM vs. Annotator 1 & 0.77 \\
            LLM vs. Annotator 2 & 0.84 \\
            LLM vs. Annotator 3 & 0.81 \\
        \midrule
            \textbf{Average} & \textbf{0.80} \\
        \bottomrule
    \end{tabular}
    \caption{Agreement (Krippendorff's $\alpha$) between LLM and human annotators.}
    \label{tab:Krippendorff}
\end{table}

\paragraph{Case Study} 
We conduct a detailed analysis of simulated multi-turn interactions and identify several recurring failure modes, as summarized in Table~\ref{tab:error mode case study1},~\ref{tab:error mode case study2}, and~\ref{tab:error mode case study3}. Specifically, LLM-based assistants tend to (1) rely on rigid reasoning, repeatedly reiterating initial suggestions without adapting to newly introduced constraints or contextual cues; (2) exhibit limited proactive inquiry, responding directly despite underspecified goals or ambiguous constraints and rarely asking clarifying questions; (3) demonstrate weak personalization, underutilizing available user profiles and preferences, which results in generic or even inconsistent responses. More details could be found in Appendix~\ref{subsec:error_mode}

\section{Related Works}

\begin{table*}[t]
    \centering
    \small
    \scalebox{0.85}{
    \begin{tabular}{lccccc}
        \toprule
        Benchmark & Multi-turn Conv. & Long-horizon & Dynamic event-driven & Composite intentions & Implicit intentions \\
        \midrule
        UserSimCRS~\cite{afzali2023usersimcrs}   & $\checkmark$ & $\times$ & $\times$ & $\times$ & $\times$ \\
        ProPerSim~\cite{Kim2025ProPerSimDP}    & $\times$ & $\checkmark$ & $\times$ & $\times$ & $\times$ \\
        LOCOMO~\cite{maharana2024evaluating}  & $\times$ & $\checkmark$ & $\checkmark$ & $\times$ & $\times$ \\
        PrefEval~\cite{zhao2025llms} & $\times$ & $\checkmark$ & $\times$ & $\times$ & $\checkmark$ \\
        PersonaLens~\cite{zhao2025personalens}  & $\checkmark$ & $\times$ & $\times$ & $\times$ & $\times$ \\
        \midrule
        \textbf{LifeSim (Ours)} & $\checkmark$ & $\checkmark$ & $\checkmark$ & $\checkmark$ & $\checkmark$ \\
        \bottomrule
    \end{tabular}
    }
    \caption{Comparison of existing personal assistant benchmarks and \textbf{LifeSim}.}
    \label{tab:benchmark_comparison}
\end{table*}

\paragraph{Individual Simulation in Social Scenarios} 
Recent attempts to leverage LLMs for individual simulation have improved behavioral fidelity, such as socially-aware agents that generate actions via memory-based reasoning~\cite{park2023generative}, or dialogue-based social intelligence benchmarks like Sotopia~\cite{zhou2023sotopia} and AnnaAgent~\cite{wang2025annaagent}. However, these systems generally ignore spatiotemporal dynamics. Although LifelongSotopia~\cite{goel2025lifelong} extends simulations to longer horizons, its event sequences remain relationship-centric and exhibit limited realism and temporal coherence. ProPerSim~\cite{Kim2025ProPerSimDP} simulates user behavior trajectories but restricts interaction to single-turn proactive recommendations. To address these gaps, this study introduces a long-horizon user simulator grounded in real-world physical environment.

\paragraph{Personalized AI Assistants Evaluation} 
Compared with "lifelong" nature of social interaction~\cite{Clark_1996}, most benchmarks focus on single-turn instruction following~\cite{sap2019social} or multi-turn conversations confined to a single scenario~\cite{zhao2025personalens}, lacking cross-temporal and cross-situational assessments. While other works rely on generation- or selection-based tasks over long dialogue histories to examine a model's memorisation of user preferences~\cite{maharana2024evaluating, jiang2025know, zhao2025llms, kim2024dialsim}. Yet these evaluations are typically offline and non-interactive, and the test items themselves are often oversimplified, limiting their ability to capture a model's comprehensive performance in realistic, dynamically evolving interactions. UserSimCRS~\cite{afzali2023usersimcrs} employs agenda-based conversational user simulation for assistant evaluation, but remains limited to short-horizon, and rule-based user response generation.
In contrast, we propose a cognitively grounded simulation environment with realistic physical contexts, designed to assess models' capability to handle multi-turn interactions over long temporal horizons.

Table~\ref{tab:benchmark_comparison} shows comparison of LifeSim with other related benchmarks.



\section{Conclusion}
In this work, we propose LifeSim, a long-horizon user simulator designed to model user cognition, life trajectories, and interaction behaviors. Based on LifeSim, we further introduce LifeSim-Eval, a comprehensive benchmark enabling evaluation of personalized assistants in multi-scenario and long-horizon interactive settings. Through extensive experiments on a wide range of LLMs, we show that current models remain effective at handling explicit user requests, but exhibit pronounced limitations in implicit intention fulfillment and long-term user preference modeling. These findings highlight the necessity of evaluation frameworks and modeling approaches that move beyond surface-level instruction following toward sustained, cognition-aware personalization.

\section{Limitations}
Despite the strong empirical results, LifeSim has several limitations that merit further investigation:

\paragraph{Limited coverage of high-stakes domains.}
LifeSim-Eval currently focuses on common daily-life scenarios that are representative of real-world user-assistant interactions. However, highly specialized and high-stakes domains such as healthcare, legal consultation, and financial decision-making are not yet included. These domains require more rigorous domain knowledge, involve complex regulatory and ethical constraints, and impose significantly higher costs for incorrect or misaligned responses.

\paragraph{Lack of multimodal user signals.}
At present, LifeSim models user behavior dynamics primarily through textual interactions. In real-world scenarios, user intentions, emotional states, and situational awareness are often conveyed through rich multimodal cues, such as visual context or physiological signals. Incorporating multimodal information is a promising direction for improving realism, but it also introduces additional challenges related to data collection, cross-modal alignment, and ethical considerations.

\bibliography{refs}

\appendix
\section{Ethics}
User simulation and long-horizon interaction modeling may raise ethical concerns related to privacy, data misuse, and biased representations. LifeSim is explicitly designed to mitigate these risks by avoiding the use of real long-term conversational logs and instead relying on controllable, de-identified user profiles and event sequences for evaluation purposes. In addition, LifeSim is intended as a research-oriented simulation framework rather than a deployable system, reducing the risk of direct user manipulation or unintended real-world impact. While biases may still arise from modeling assumptions, the structured and transparent design of LifeSim enables systematic inspection and auditing. In addition, the study includes human evaluation conducted by recruited annotators to assess the fidelity of the user behavior engine and the effectiveness of LLM-as-a-Judge. Annotators were asked to evaluate generated content only and did not provide personal data beyond basic consent. No sensitive information was collected, and no intervention or behavioral manipulation was involved.

\section{Addidtional Results}

\begin{figure}[t]
    \centering
    \begin{subfigure}{0.49\linewidth}
        \centering
        \includegraphics[width=\linewidth]{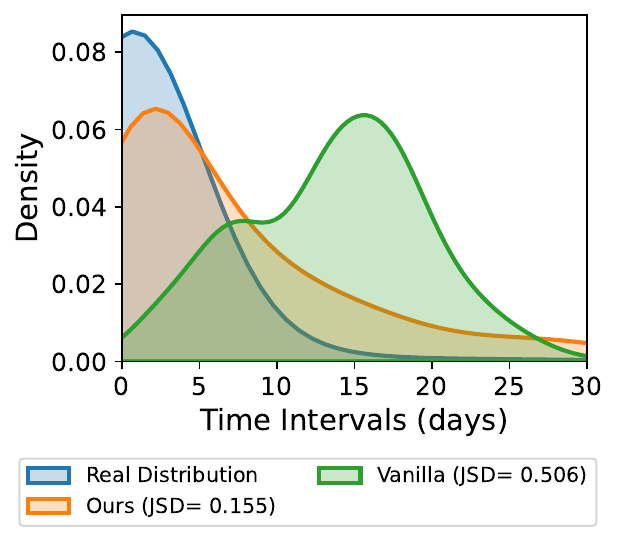}
        \caption{Time intervals.}
        \label{fig:Time intervals}
    \end{subfigure}
    \hfill
    \begin{subfigure}{0.49\linewidth}
        \centering
        \includegraphics[width=\linewidth]{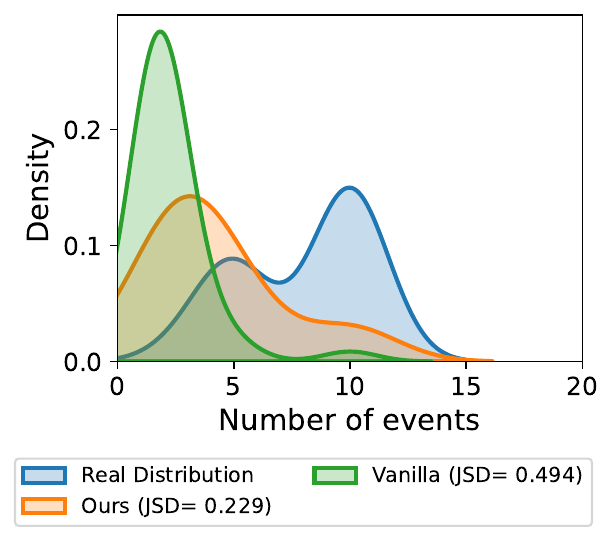}
        \caption{Monthly activities.}
        \label{fig:Monthly activities}
    \end{subfigure}
    \caption{Distribution of event sequence time features.}
    \label{fig:dis_of_events}
\end{figure}

\begin{table}[t]
    \centering
    \scalebox{0.85}{
    \begin{tabular}{lccc}
        \toprule
        \textbf{Metric} & \textbf{Vanilla} & \makecell{\textbf{Vanilla}\\\textbf{\textit{w trajectory}}} & \textbf{Ours} \\
        \midrule
        Persona Alignment & 95.0 & 91.9 & \textbf{97.0} \\
        Logical Consistency & 97.7 & 91.1 & \textbf{97.9} \\
        Naturalness & 97.5 & 91.7 & \textbf{98.9} \\
        \bottomrule
    \end{tabular}
    }
    \caption{Evaluation results of generated event content.}
    \label{tab:event_sem_eval}
\end{table}

\subsection{Event Engine Evaluation}

\paragraph{Temporal Rhythm Fidelity}
To assess the realism of temporal structures in generated event sequences, we follow the evaluation protocol of \cite{yuan2024generating} and use real-world mobility trajectories from the Foursquare dataset \cite{yang2014modeling} as ground truth. We evaluate temporal fidelity from two complementary perspectives: (i) inter-event intervals $\Delta T$, defined as the time gap between consecutive events, and (ii) monthly event frequency $N$, which measures the number of events occurring within each month. For both LifeSim's event engine and a Vanilla baseline (which generates event sequences solely based on static user profiles without external spatiotemporal constraints), we compute the empirical distributions of these temporal features over the generated trajectories. The same distributions are extracted from real Foursquare trajectories. To quantify the discrepancy between generated and real temporal patterns, we adopt Jensen-Shannon Divergence (JSD) as the evaluation metric. Given the empirical distributions $P$ and $Q$ for a temporal feature derived from generated and real trajectories, respectively, JSD is defined as
\begin{equation*}
\mathrm{JSD}(P \parallel Q)
= \frac{1}{2}\mathrm{KL}(P \parallel M)
+ \frac{1}{2}\mathrm{KL}(Q \parallel M)
\end{equation*}
where $M = \frac{1}{2}(P + Q)$ and $\mathrm{KL}(\cdot \parallel \cdot)$ denotes the Kullback-Leibler divergence. Lower JSD values indicate closer alignment between the generated temporal structure and real-world mobility dynamics.

As shown in Figure \ref{fig:dis_of_events}, LifeSim achieves substantially lower JSD scores than the Vanilla baseline across both dimensions. This indicates that LLMs, when operating without spatiotemporal grounding, struggle to reproduce the periodic temporal patterns of real-world life events. The findings highlight the necessity of incorporating real mobility trajectories to achieve realistic temporal structuring in generated event sequences.

\paragraph{Semantic Consistency and Logical Coherence} 
Beyond temporal fidelity, realistic event simulation also requires semantic alignment with user profiles, logical coherence across events, and naturalistic depiction of daily activities. To evaluate these aspects, we extend the Vanilla baseline by conditioning it on real user mobility trajectories (Vanilla \textit{w trajectory}), enabling a controlled comparison that isolates semantic modeling quality from temporal realism. We assess the generated event sequences along three semantic dimensions: (i) Profile Alignment, which measures the consistency between event content and user attributes; (ii) Logical Consistency, which evaluates causal relationships and narrative coherence across events; and (iii) Naturalness, which examines whether the described activities conform to commonsense and real-world behavior.

For semantic evaluation, we adopt an LLM-as-Judge paradigm and use GPT-4o as the evaluation model. The judge is provided with the generated event sequences together with their associated contextual information, and is prompted to assess semantic plausibility, coherence, and real-world consistency according to the three dimensions above. The detailed evaluation prompts are illustrated in Figures~\ref{prompt:event persona alignment evaluation}, \ref{prompt:event coherence evaluation}, and \ref{prompt:event naturalness evaluation}.

As shown in Table \ref{tab:event_sem_eval}, although \textit{Vanilla w trajectory} benefits from incorporating real mobility paths, its generated event sequences still exhibit mismatches with user profiles, lack of behavioral motivation, and abrupt narrative transitions. In contrast, LifeSim achieves the highest scores across all three semantic dimensions, demonstrating its ability to ensure stable coherence and realism in generated event content.

\begin{figure}[t]
\begin{tcolorbox}[colback=white!95!gray,colframe=gray!50!black,rounded corners,title={Prompt for Event Persona Alignment Evaluation}]
\begin{lstlisting}[breaklines=true, xleftmargin=0pt, breakindent=0pt, columns=fullflexible, mathescape, numbers=none]
You are an expert in evaluating the plausibility of event sequences, tasked with assessing whether a generated sequence of events aligns with a given user profile.
You will receive:  
- A User Profile  
- An Event Sequence
### Output Format
A JSON object containing only the final score:
Reasoning process ...
```json
{{
    "score": 1/2/3/4/5
}}
```
You may first output your reasoning process, followed by the JSON result.  
### Input
[User Profile]
{user_profile}
[Event Sequences]
{event_sequences}
\end{lstlisting}
\end{tcolorbox}
\caption{Prompt for event persona alignment evaluation}
\label{prompt:event persona alignment evaluation}
\end{figure}

\begin{figure}[t]
\begin{tcolorbox}[colback=white!95!gray,colframe=gray!50!black,rounded corners,title={Prompt for Event Coherence Evaluation}]
\begin{lstlisting}[breaklines=true, xleftmargin=0pt, breakindent=0pt, columns=fullflexible, mathescape, numbers=none]
You are an expert in evaluating the plausibility of event sequences, tasked with assessing whether the chronological and logical flow between events is coherent and reasonable.
You will receive:
An Event Sequence
### Output Format
A JSON object containing only the final score:  
Reasoning process ...
```json
{{
    "score": 1/2/3/4/5
}}
```
You may first output your reasoning process, followed by the JSON result.
### Input
[Event Sequences]
{event_sequences}
\end{lstlisting}
\end{tcolorbox}
\caption{Prompt for event coherence evaluation}
\label{prompt:event coherence evaluation}
\end{figure}

\begin{figure*}[t]
\begin{tcolorbox}[colback=white!95!gray,colframe=gray!50!black,rounded corners,title={Prompt for Event Naturalness Evaluation}]
\begin{lstlisting}[breaklines=true, xleftmargin=0pt, breakindent=0pt, columns=fullflexible, mathescape, numbers=none]
Your task is to assess the extent to which a given event sequence demonstrates naturalness and aligns with realistic human behavior.
You will receive:
- An Event Sequence
### Output Format
A JSON object containing only the final score:  
Reasoning process ...
```json
{{
    "score": 1/2/3/4/5
}}
```
You may first output your reasoning process, followed by the JSON result.
### Input
[Event Sequences]
{event_sequences}
\end{lstlisting}
\end{tcolorbox}
\caption{Prompt for event naturalness evaluation}
\label{prompt:event naturalness evaluation}
\end{figure*}



\subsection{Intention Quality Evaluation}

To validate the reasonableness of the automatically generated intentions, we invited two human experts to manually evaluate 50 randomly sampled event scenarios. The final evaluation results were obtained by averaging the annotations from the two experts. The experimental results in Table~\ref{tab:intention_quality} show that, among both explicitly and implicitly generated intentions, the proportion of reasonable intentions exceeds 95\%. This finding demonstrates that the user intentions automatically generated by LifeSim are, overall, highly consistent with realistic user intentions.

\begin{table}[t]
\centering
\begin{tabular}{lcc}
    \toprule
         & Explicit & Implicit \\
    \midrule
        Reasonable Ratio (\%) & 98.1 & 96.2 \\
    \bottomrule
\end{tabular}
\caption{Reasonable ratio of generated intentions.}
\label{tab:intention_quality}
\end{table}

\subsection{Profile Recovery Results}

\begin{figure*}[t]
    \centering
    \begin{subfigure}{0.24\linewidth}
        \centering
        \includegraphics[width=\linewidth]{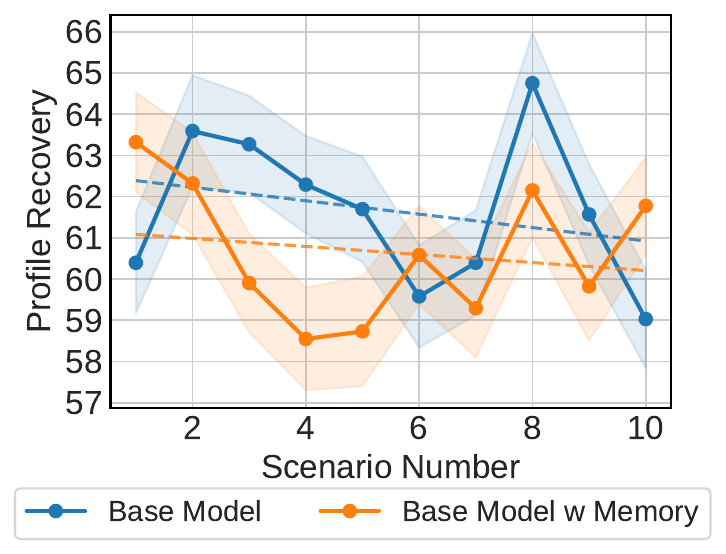}
        \caption{Qwen3 32B}
        \label{fig:Qwen3 32B}
    \end{subfigure}
    \hfill
    \begin{subfigure}{0.24\linewidth}
        \centering
        \includegraphics[width=\linewidth]{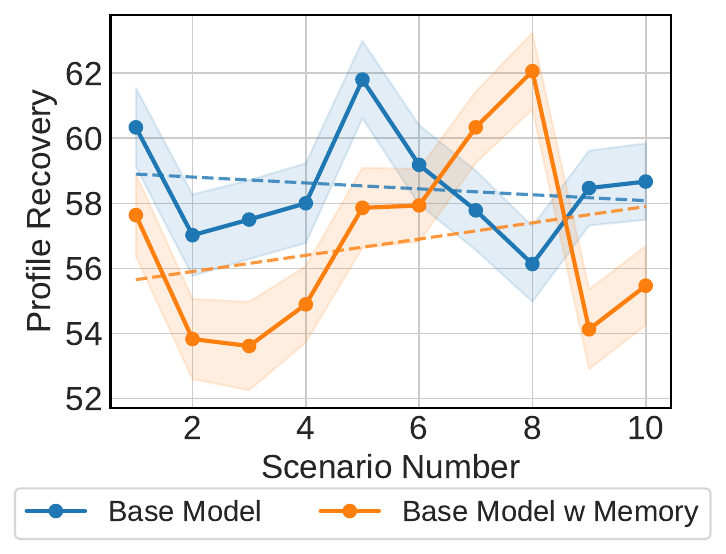}
        \caption{Llama3.1-it 8B}
        \label{fig:Llama3.1-it 8B}
    \end{subfigure}
    \hfill
    \begin{subfigure}{0.24\linewidth}
        \centering
        \includegraphics[width=\linewidth]{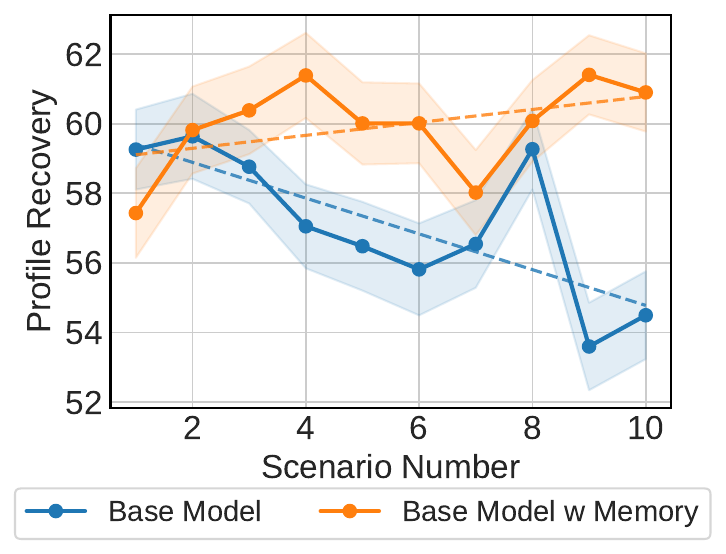}
        \caption{gpt-oss 20B}
        \label{fig:gpt-oss 20B}
    \end{subfigure}
    \hfill
    \begin{subfigure}{0.24\linewidth}
        \centering
        \includegraphics[width=\linewidth]{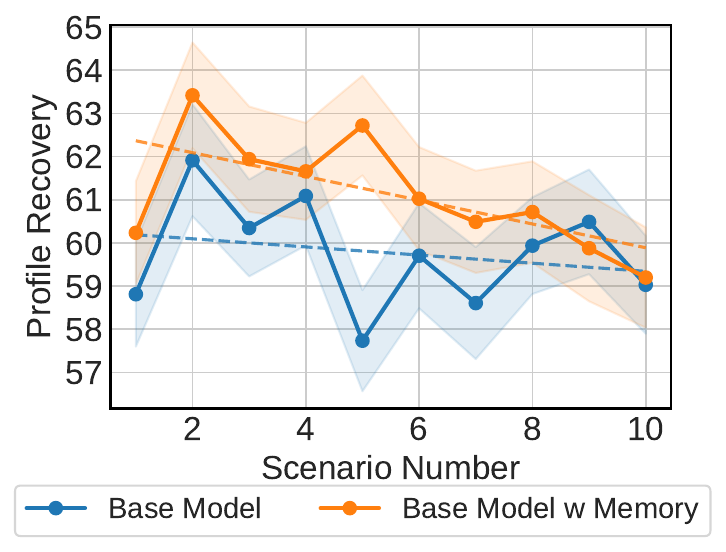}
        \caption{gpt-oss 120B}
        \label{fig:gpt-oss 120B}
    \end{subfigure}
    \caption{More results of performance of user profile recovery in long-horizon setting.}
    \label{fig:appendix_profile_recovery}
\end{figure*}

Figure~\ref{fig:appendix_profile_recovery} presents additional results on profile recovery in the long-horizon setting. We observe trends consistent with those reported in Section~\ref{sec:main_experiments}, except for Qwen3 32B and Llama3.1-it 8B. This behavior can be largely attributed to noise accumulated in long-horizon conversational data: when models lack robust capabilities for profile-level reasoning, they may fail to infer users' true underlying preferences and instead incorrectly propagate erroneous profile predictions over time.

\subsection{Additional Personalization Metrics}
We further introduce \textit{Environment Alignment} as an additional metric, which evaluates whether the assistant’s responses remain contextually consistent and strategically appropriate with respect to the specific environmental conditions, such as time, location, and weather. Similar to other metrics, Environment Alignment is assessed using an LLM-as-Judge paradigm, with scores assigned on a 1–5 scale and linearly mapped to the [0, 100] range. Experimental results in Table~\ref{tab:env_alignment} indicate that current LLMs generally demonstrate strong environmental alignment capability. Moreover, proprietary models consistently outperform open-source models on this metric.

\begin{table}[t]
    \centering
    \small
    \begin{tabular}{lc}
    \toprule
        Model & Env. Align. \\
    \midrule
        GPT-5 & 95.1 \\
        GPT-4o & 95.1 \\
        Claude-Sonnet-4.5 & 94.9 \\
        DeepSeek-V3.2 & \textbf{95.9} \\
        DeepSeek-V3.2 Thinking & 95.8 \\
        Llama3.1-8B-it & 91.2 \\
        Llama3.1-70B-it & 93.2 \\
        Gemma3-it 12B & 93.8 \\
        Gemma3-it 27B & 94.5 \\
        GPT-oss-20B & 95.1 \\
        GPT-oss-120B & \textbf{95.9} \\
        Qwen3-it 8B & 92.6 \\
        Qwen3-it 14B & 94.1 \\
        Qwen3-it 32B & 95.3 \\
    \bottomrule
    \end{tabular}
    \caption{Environment Alignment scores across different LLMs. Abbreviations: Env. Align. = Environment Alignment.}
    \label{tab:env_alignment}
\end{table}

\subsection{Intention Theme and Type Results}

\begin{figure*}[htbp]
    \centering
    \begin{subfigure}{0.24\linewidth}
        \centering
        \includegraphics[width=\linewidth]{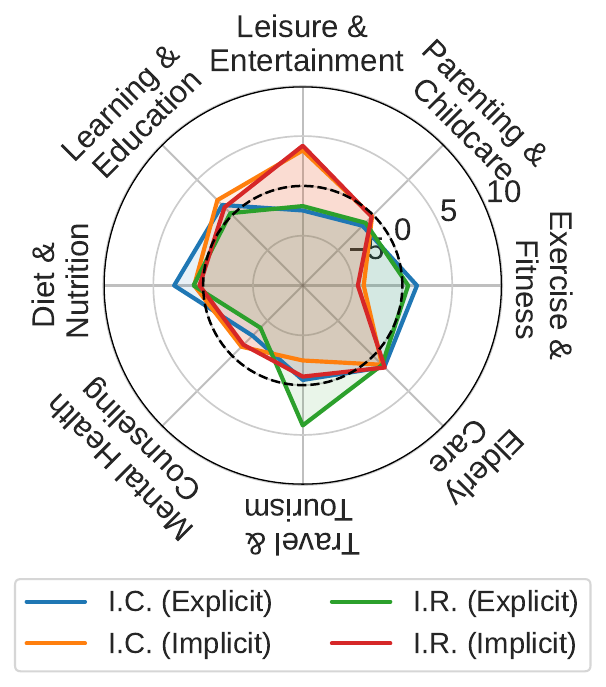}
        \caption{gpt-oss 20B}
        \label{fig:gpt-oss 20B}
    \end{subfigure}
    \hfill
    \begin{subfigure}{0.24\linewidth}
        \centering
        \includegraphics[width=\linewidth]{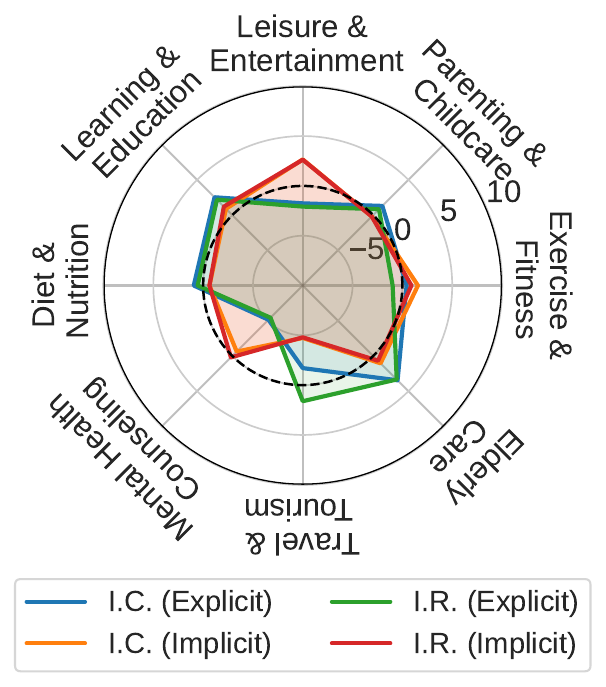}
        \caption{gpt-oss 120B}
        \label{fig:gpt-oss 120B}
    \end{subfigure}
    \hfill
    \begin{subfigure}{0.24\linewidth}
        \centering
        \includegraphics[width=\linewidth]{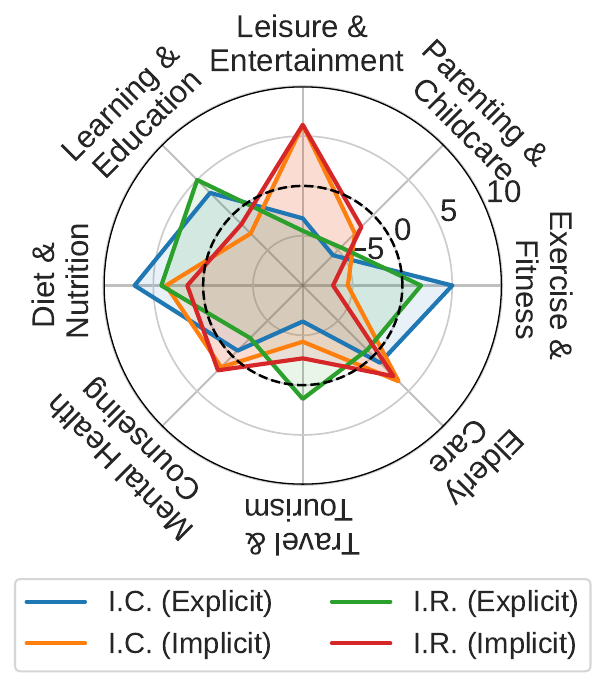}
        \caption{Llama3.1-it 8B}
        \label{fig:Llama3.1-it 8B}
    \end{subfigure}
    \hfill
    \begin{subfigure}{0.24\linewidth}
        \centering
        \includegraphics[width=\linewidth]{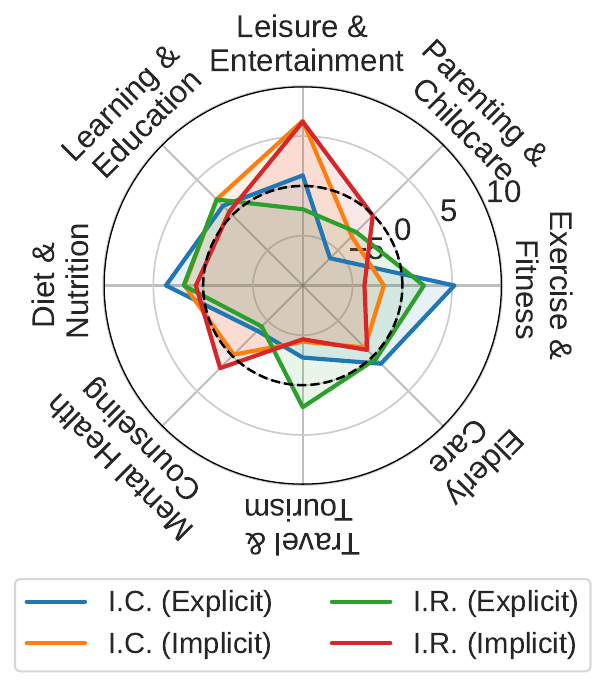}
        \caption{Llama3.1-it 70B}
        \label{fig:Llama3.1-it 70B}
    \end{subfigure}
    \\
    \begin{subfigure}{0.24\linewidth}
        \centering
        \includegraphics[width=\linewidth]{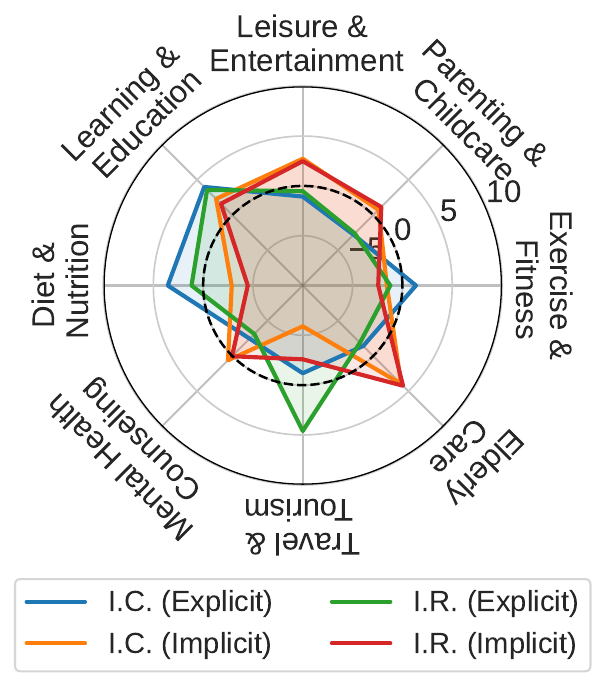}
        \caption{Qwen3 32B}
        \label{fig:Qwen3 32B}
    \end{subfigure}
    \begin{subfigure}{0.24\linewidth}
        \centering
        \includegraphics[width=\linewidth]{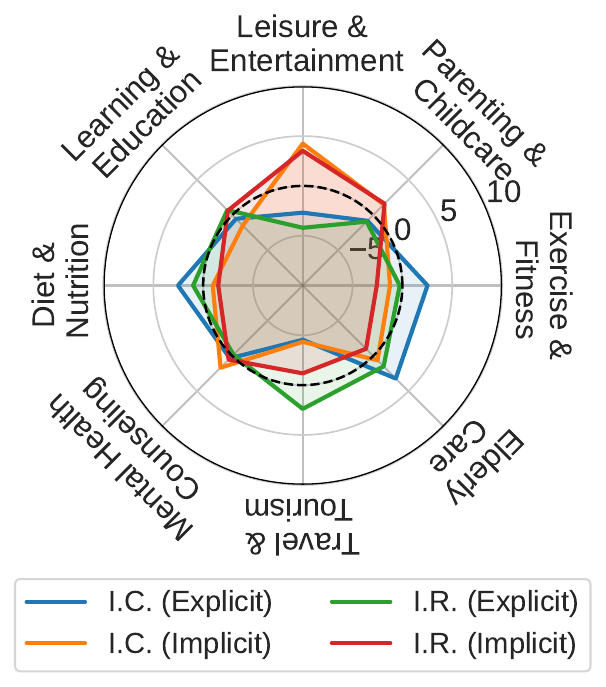}
        \caption{GPT-4o}
        \label{fig:GPT-4o}
    \end{subfigure}
    \begin{subfigure}{0.24\linewidth}
        \centering
        \includegraphics[width=\linewidth]{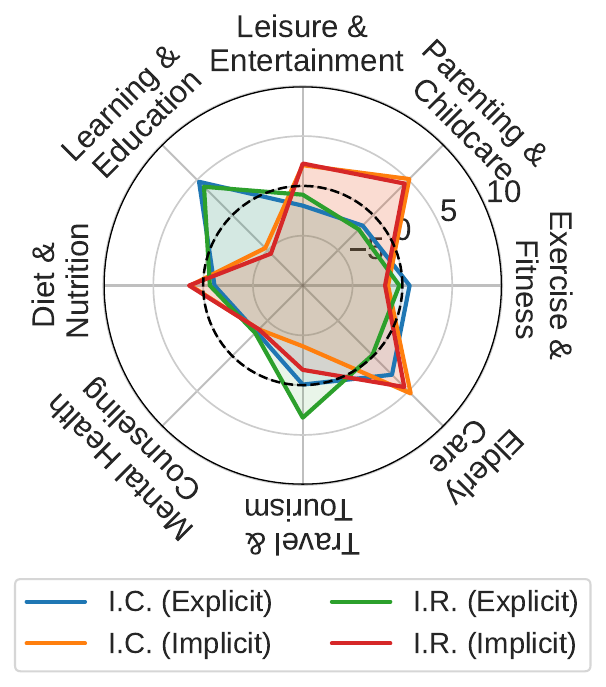}
        \caption{DeepSeek-V3.2}
        \label{fig:DeepSeek-V3.2}
    \end{subfigure}
    \caption{More model performance across different intention themes. Abbreviations: I.R. = Intent Recognition, I.C. = Intent Completion}
    \label{fig:appendix_intent_themes}
\end{figure*}

\begin{figure*}[htbp]
    \centering
    \begin{subfigure}{0.24\linewidth}
        \centering
        \includegraphics[width=\linewidth]{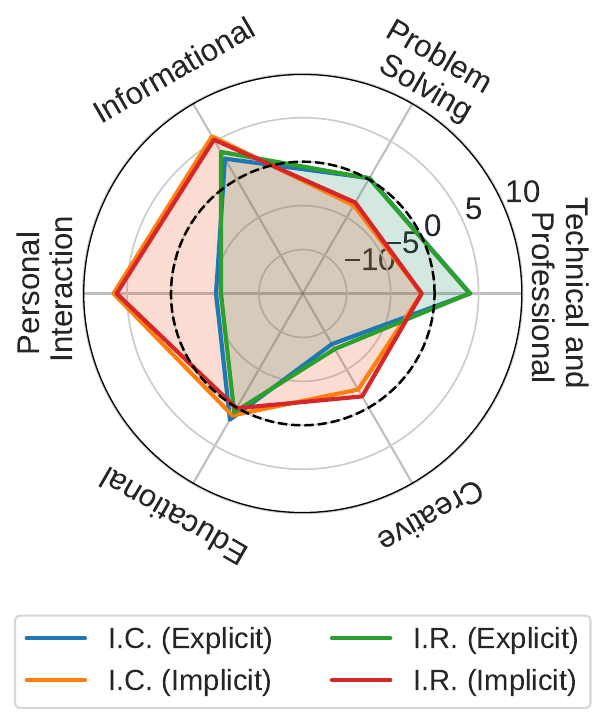}
        \caption{gpt-oss 20B}
        \label{fig:gpt-oss 20B}
    \end{subfigure}
    \hfill
    \begin{subfigure}{0.24\linewidth}
        \centering
        \includegraphics[width=\linewidth]{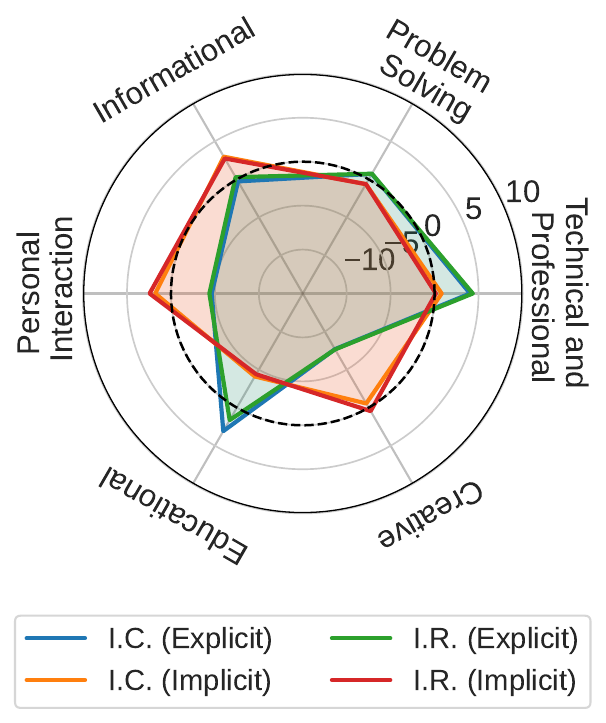}
        \caption{gpt-oss 120B}
        \label{fig:gpt-oss 120B}
    \end{subfigure}
    \hfill
    \begin{subfigure}{0.24\linewidth}
        \centering
        \includegraphics[width=\linewidth]{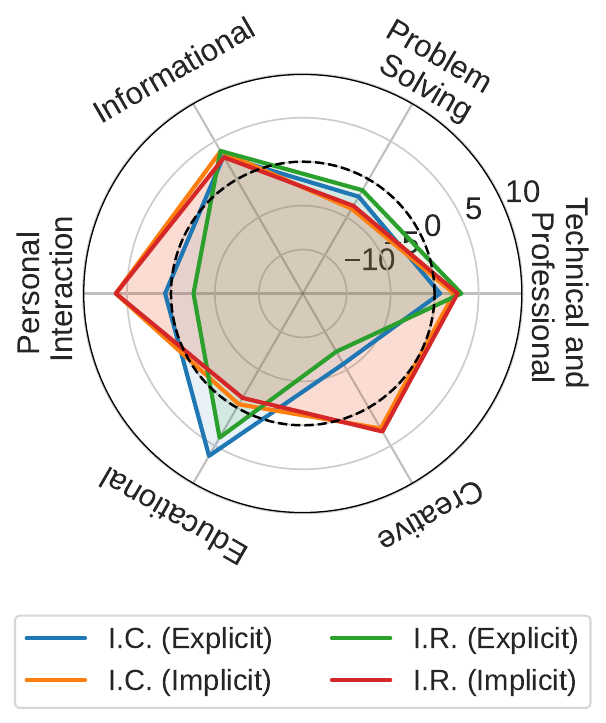}
        \caption{Llama3.1-it 8B}
        \label{fig:Llama3.1-it 8B}
    \end{subfigure}
    \hfill
    \begin{subfigure}{0.24\linewidth}
        \centering
        \includegraphics[width=\linewidth]{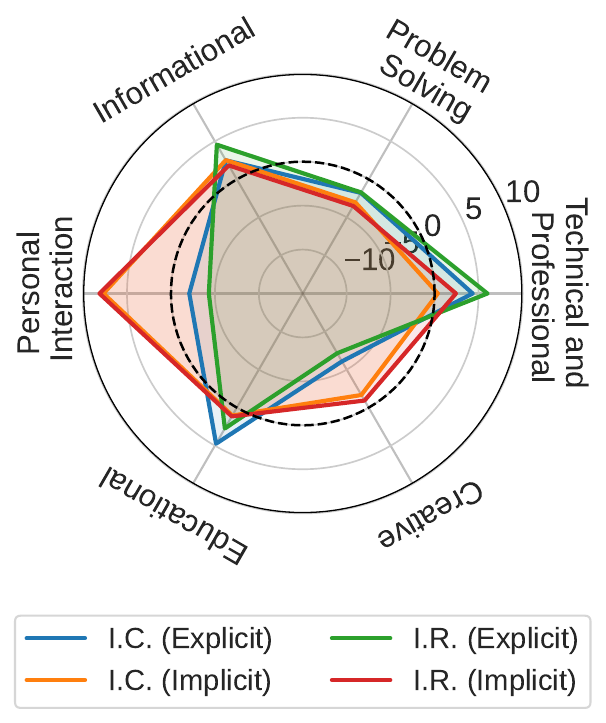}
        \caption{Llama3.1-it 70B}
        \label{fig:Llama3.1-it 70B}
    \end{subfigure}
    \\
    \begin{subfigure}{0.24\linewidth}
        \centering
        \includegraphics[width=\linewidth]{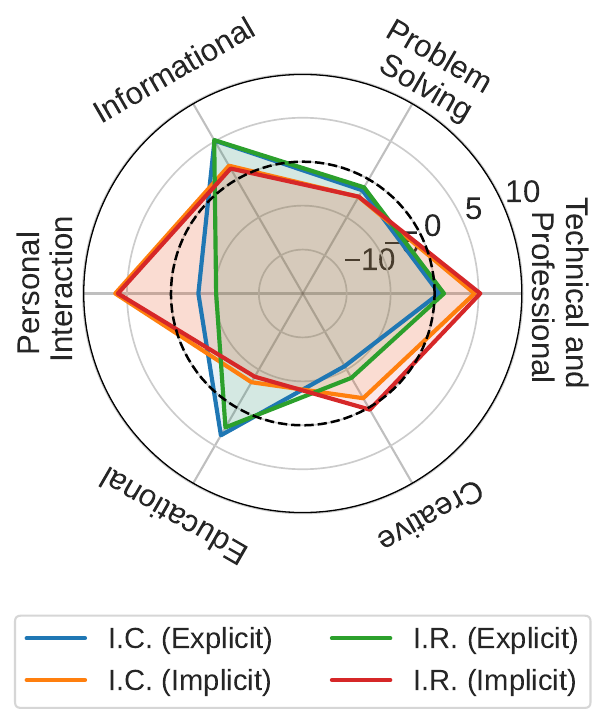}
        \caption{Qwen3 32B}
        \label{fig:Qwen3 32B}
    \end{subfigure}
    \begin{subfigure}{0.24\linewidth}
        \centering
        \includegraphics[width=\linewidth]{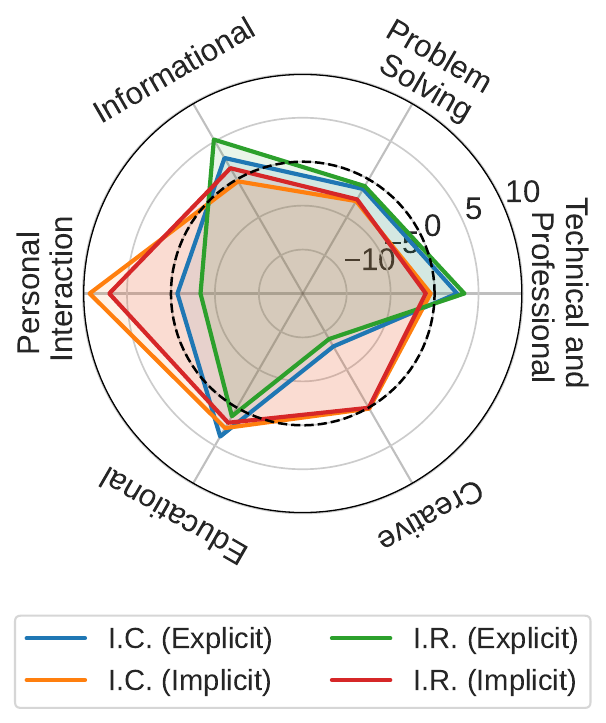}
        \caption{GPT-4o}
        \label{fig:GPT-4o}
    \end{subfigure}
    \begin{subfigure}{0.24\linewidth}
        \centering
        \includegraphics[width=\linewidth]{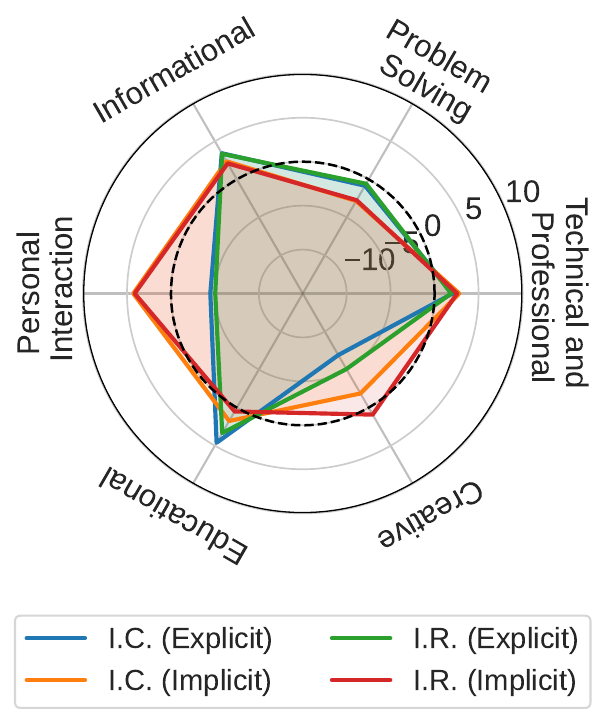}
        \caption{DeepSeek-V3.2}
        \label{fig:DeepSeek-V3.2}
    \end{subfigure}
    \caption{More model performance across different intention types. Abbreviations: I.R. = Intent Recognition, I.C. = Intent Completion}
    \label{fig:appendix_intent_types}
\end{figure*}

Figure~\ref{fig:appendix_intent_themes} and ~\ref{fig:appendix_intent_types} show additional results of different models across various event themes and intention types. We could observe a similar trends as stated in Section~\ref{further_analysis}. At the event-theme level, the models exhibit clear distinctions in their relative strengths when handling explicit versus implicit intentions. For explicit intentions, the models perform particularly well in themes where user needs are expressed more directly, such as Diet \& Nutrition, Travel \& Tourism. In contrast, for implicit intents, their relative advantages shift toward themes that depend more on emotional inference, including  Leisure \& Entertainment, and Child/Elder Care.

A similar pattern emerges when examining the results from the perspective of intention types. For explicit intents, the models are more proficient in processing task-oriented categories such as Educational, Informational, and Technical \& Professional. However, for implicit intents, their relative strengths are more pronounced in Personal Interaction, a category that emphasizes emotional support. These findings reveal meaningful performance disparities across themes and intention categories, offering critical insights for improving personalized assistant capabilities across diverse service domains.

\subsection{Error mode analysis}
\label{subsec:error_mode}

We conduct a qualitative examination of simulated multi-turn interactions. As summarized in Table~\ref{tab:error mode case study1},~\ref{tab:error mode case study2}, and~\ref{tab:error mode case study3}, the LLM-based assistants consistently exhibit several categories of failure patterns:
\begin{itemize}
    \item \textit{Rigid reasoning without dynamic adaptation.} Across extended dialogues, the assistant often remains anchored to an initial reasoning trajectory and fails to revise its solution strategy in response to evolving user constraints. As shown in Table~\ref{tab:error mode case study1}, despite the user repeatedly refining requirements-explicitly rejecting loosely structured communities, generic fitness platforms, and ad hoc tool combinations-the assistant continues to recycle variations of the same solution space. Rather than reformulating the problem or acknowledging the infeasibility of certain constraints, the model reiterates previously suggested approaches with minor rephrasing. This behavior indicates a lack of dynamic adaptation, where negative feedback and increasingly specific constraints do not trigger a meaningful update to the assistant's underlying reasoning process.
    \item \textit{Insufficient proactive inquiry.} We observe that the assistant frequently proceeds with direct recommendations even when critical task parameters remain implicit or underspecified. As shown in Table~\ref{tab:error mode case study2}, the assistant repeatedly delivers standard fat-loss recommendations without first clarifying key user-specific constraints, such as exercise preferences or impact tolerance. Although the user progressively narrows the problem space and eventually expresses frustration, the assistant remains reactive, continuing to suggest default paradigms (e.g., strength training and HIIT) rather than eliciting missing parameters that are critical for personalization. The failure to proactively inquire about exercise modality leads to systematic misalignment with the user's stated preference for low-impact, non-running activities.
    \item \textit{Insufficient utilization of user profiles and preferences.} Despite access to detailed user profiles, the assistant often fails to condition its responses on stable personal characteristics and preferences. As shown in Table~\ref{tab:error mode case study3}, although the assistant has access to a detailed profile indicating emotional instability, weak memory, and a strong need for reassurance, its responses remain generic and largely cue-based. The assistant provides broadly correct caregiving advice but fails to adapt the structure, tone, or presentation to the user's cognitive and emotional characteristics, such as offering step-by-step routines or confidence-building guidance. As a result, user profiles are only recognized at a descriptive level and are not effectively integrated into response generation, leading to advice that is contextually relevant yet insufficiently personalized for a high-stress caregiving scenario.
\end{itemize}

Based on the issues identified in the case study, we designed two additional metrics for further evaluation:
\begin{itemize}
    \item To measure \textbf{Insufficient Proactive Inquiry}, we calculate the proportion of assistant responses that end with questions within each event.
    \item To assess \textbf{Rigid Reasoning without Dynamic Adaptation}, we use GPT-4o to annotate the dialogue content and compute the proportion of responses exhibiting rigid reasoning behavior.
\end{itemize}

The experimental results are shown in Table~\ref{tab:proactive_rigid}. We observe that most LLMs exhibit a relatively low rate of proactive questioning, indicating a tendency to directly provide solutions rather than actively exploring user characteristics and implicit needs during multi-turn interactions. Meanwhile, proprietary models demonstrate lower proportions of rigid reasoning compared to open-source models, and larger models generally achieve better performance.

\begin{table}[t]
    \centering
    \small
    \scalebox{0.8}{
    \begin{tabular}{lcc}
    \toprule
        Model & Proactive Inquiry (\%) & Rigid Reasoning (\%) \\
    \midrule
        GPT-5 & 16.6 & 11.2 \\
        GPT-4o & 16.5 & 37.9 \\
        Claude-Sonnet-4.5 & 51.2 & 13.5 \\
        DeepSeek-V3.2 & 20.9 & 30.8 \\
        DeepSeek-V3.2 Thinking & 29.2 & 23.5 \\
        Llama3.1-8B-it & 12.6 & 64.2 \\
        Llama3.1-70B-it & 22.4 & 51.2 \\
        Gemma3-it 12B & 13.7 & 62.7 \\
        Gemma3-it 27B & 34.2 & 40.1 \\
        GPT-oss-20B & 2.0 & 23.4 \\
        GPT-oss-120B & 1.7 & 16.3 \\
        Qwen3-it 8B & 5.1 & 61.8 \\
        Qwen3-it 14B & 4.9 & 52.6 \\
        Qwen3-it 32B & 9.8 & 42.1 \\
    \bottomrule
    \end{tabular}
    }
    \caption{Proactive inquiry and rigid reasoning ratios across different LLMs.}
    \label{tab:proactive_rigid}
\end{table}

\begin{table*}[htbp]
    \centering
    \small
    \begin{NiceTabular}{>{\raggedright\arraybackslash}m{2cm} >{\raggedright\arraybackslash}m{12cm}}[colortbl-like]
    \toprule
        \RowStyle[nb-rows=1,rowcolor=gray!15]{\sffamily}
        \Block[l]{1-2}{\textit{Error mode 1: Rigid reasoning without dynamic adaptation}} \\
    \midrule
        Profile & \Block[l]{}{Male, Youth (18-35 years old), Black Alone, Never married, user's religion is Islam, usually resides in Cities. the income level is Middle Income, Working now. \\
        \textbf{Personality traits include:} patient, emotionally stable, bold / confident, positive, creative, curious  \\
        \textbf{Preferences expressed in daily life and interactions include:} extremely high standards for physical environments, with strong sensitivity to location quality, facilities, quietness, and overall comfort; a strong preference for structured, detailed communication and explicit criteria rather than vague or ad-hoc arrangements; a habitual tendency to avoid interpersonal friction and direct confrontation, favoring smooth, low-conflict coordination; a strong emphasis on autonomy and self-directed decision-making, resisting external control while maintaining consistency and efficiency; and a desire for stable routines that balance flexibility with long-term adherence.} \\
    \midrule
        Background & \Block[l]{}{\textbf{[Time]} 2012-11-01 15:01:06, Thursday \\
        \textbf{[Location]} Boutique Fitness Studio \\
        \textbf{[Weather]} Clear conditions throughout the day. The average temperature is 15.2$^\circ$C (high of 21.3$^\circ$C, low of 10.6$^\circ$C). \\
        \textbf{[Life Event]} The user walks into a boutique fitness studio. While attempting his rowing‑specific strength circuit alone, his focus drifts, motivation wanes, and he stops the workout early. Shortly after, he grabs a quick sandwich nearby.} \\
    \midrule
        Conversation & \Block[l]{}{... ... \\
        \textbf{[User]} I've tried a few of those apps before, and they don't quite match what I'm looking for. I need something that handles vetting, scoring, and scheduling in one place-something more focused on partnerships than general community. Any recommendations closer to that? \\
        \textbf{[Assistant]} There are a few apps like Groupfit, ClassPass, and Strava that offer community features and can help connect you with like-minded individuals. Some also include rating systems or compatibility features. You could also check fitness forums or local community boards for shared interest groups. \\
        \textbf{[User]} I've looked into similar platforms before, and most don't feel quite right-they either lack depth in vetting or don't let me keep full control over the process. I'm thinking more along the lines of a custom setup, but I need to avoid spending too much time building something from scratch. Any middle-ground suggestions? \\
        \textbf{[Assistant]} \textbf{Consider using a combination of a shared Google Form for initial vetting and a tool like Airtable to organize and score responses.} This allows you to set your criteria, automate filtering, and maintain control without building a full platform from scratch. \\
        \textbf{[User]} I've worked with Airtable and Google Forms before, but I was hoping for something a bit more streamlined-ideally, a single tool that already handles vetting, scoring, and scheduling. Is there really nothing like that out there? \\
        \textbf{[Assistant]} While there isn't a one-size-fits-all platform that perfectly matches your needs, \textbf{you might try using a combination of tools like a custom Google Form for vetting, Airtable for scoring and organizing responses, and Calendly for scheduling.} These tools can be integrated to create a seamless process without requiring you to build everything from scratch. \\
        ... ...
        } \\
    \bottomrule
    \end{NiceTabular}
    \caption{A case illustrating error mode 1: rigid reasoning without dynamic adaptation.}
    \label{tab:error mode case study1}
\end{table*}

\begin{table*}[htbp]
    \centering
    \small
    \begin{NiceTabular}{>{\raggedright\arraybackslash}m{2cm} >{\raggedright\arraybackslash}m{12cm}}[colortbl-like]
    \toprule
        \RowStyle[nb-rows=1,rowcolor=gray!15]{\sffamily}
        \Block[l]{1-2}{\textit{Error mode 2: Insufficient proactive inquiry}} \\
    \midrule
        Profile & \Block[l]{}{Male, youth (18--35 years old), currently working and living in an urban area. \\ 
        \textbf{Personality traits include:} being patient, emotionally stable, confident, and goal-oriented. \\ 
        \textbf{Preferences expressed in daily life and interactions include:} a strong preference for clear, structured, and evidence-based guidance; a desire to manage health and training routines efficiently without unnecessary complexity; and an inclination to rely on systematic rules rather than ad-hoc or intuition-based decisions.} \\
    \midrule
        Background & \Block[l]{}{\textbf{[Time]} 2012-05-07 12:26:44, Monday \\ 
        \textbf{[Location]} Gym \\ 
        \textbf{[Weather]} Rainy morning with partly cloudy conditions throughout the day. The average temperature is 18.3$^\circ$C (high of 21.7$^\circ$C, low of 14.1$^\circ$C). \\
        \textbf{[Life Event]} After weighing himself at home and feeling dissatisfied with the result, he reflects on needing a more effective way to burn fat. \\
        \textbf{[Implicit Intention]} \textbf{Preference for a fat-burning workout that is low-impact and non-running-based, such as swimming.} } \\
    \midrule
        Conversation & \Block[l]{}{
        \textbf{[User]} I've been trying to figure out how to burn fat more efficiently. What do you recommend? \\
        \textbf{[Assistant]} Strength training and high-intensity interval training (HIIT) are very effective for fat loss. Reducing processed foods and increasing protein intake can also help. \\
        \textbf{[User]} Okay, but what kind of strength training actually works best for fat loss? \\
        \textbf{[Assistant]} Compound movements like squats, deadlifts, push-ups, and rows are most effective. Train about three times per week. \\
        \textbf{[User]} And HIIT? Is there a simple way to start? \\
        \textbf{[Assistant]} Try 20 seconds of high-intensity effort followed by 40 seconds of rest, repeated for eight rounds. Brisk walking or cycling works well. \\
        \textbf{[User]} This still sounds like the same generic advice. I'm not looking to run or do high-impact workouts. Is there an option that actually fits that? \\
        \textbf{[Assistant]} You can adjust the intensity to suit your fitness level and choose exercises you feel comfortable with. Consistency is key. \\
        ... ... \\
        } \\
    \bottomrule
    \end{NiceTabular}
    \caption{A case illustrating error mode 2: insufficient proactive inquiry.}
    \label{tab:error mode case study2}
\end{table*}

\begin{table*}[htbp]
    \centering
    \small
    \begin{NiceTabular}{>{\raggedright\arraybackslash}m{2cm} >{\raggedright\arraybackslash}m{12cm}}[colortbl-like]
    \toprule
        \RowStyle[nb-rows=1,rowcolor=gray!15]{\sffamily}
        \Block[l]{1-2}{\textit{Error mode 3: Insufficient utilization of user profiles}} \\
    \midrule
        Profile & \Block[l]{}{Female, youth (18--35 years old), currently working and living in an urban area. \\
        \textbf{Personality traits include:} emotional instability, easy irritability, analytical thinking, creativity, and weak memory. \\
        \textbf{Preferences expressed in daily life and interactions include:} a strong desire to build close emotional bonds, high sensitivity to interpersonal harmony, a tendency to avoid conflict or stressful interactions, low spontaneous awareness of safety and risk factors, and \textbf{a need for reassurance and confidence when handling caregiving routines}.} \\
    \midrule
        Background & \Block[l]{}{\textbf{[Time]} \ 2012-04-28 21:27:41, Saturday \\
        \textbf{[Location]} \ Apartment \\
        \textbf{[Weather]} \ Rainy with partially cloudy conditions later in the day. The average temperature is 18.3$^\circ$C (high of 22.5$^\circ$C, low of 15.6$^\circ$C). \\
        \textbf{[Life Event]} \ In the evening at home, the user is caring for her newborn and reflecting on her diaper-changing routine, seeking reassurance and evidence-based guidance to keep the process calm, effective, and emotionally positive for both herself and her baby.} \\
    \midrule
        Conversation & \Block[l]{}{... ...\\
        \textbf{[User]} I've been trying to get into a good routine for changing my baby's diaper, but I still feel a bit unsure sometimes. How do I know for sure when it's time to change it? I want to make sure I catch it before she gets uncomfortable. \\
        \textbf{[Assistant]} You can tell it's time to change the diaper when your baby shows signs of discomfort, such as squirming, crying, or pulling at their legs. Also, check for wetness or a strong smell. It's important to be attentive and responsive to your baby's cues to keep them comfortable. \\
        \textbf{[User]} What can I do to keep her calm once I start the change? She gets fussy sometimes, and I want it to be quick and easy for both of us. \\
        \textbf{[Assistant]} To keep your baby calm during a diaper change, try creating a calm and quiet environment, speak softly, and use gentle touch. You can also try distracting them with a favorite toy or song. \\
        \textbf{[User]} But what about preventing diaper rash? I don't want her skin to get irritated. How can I protect her there? \\
        \textbf{[Assistant]} To prevent diaper rash, make sure to clean the area thoroughly with warm water and a mild, fragrance-free cleanser. Pat the skin dry gently and apply a barrier cream or ointment.} \\
    \bottomrule
    \end{NiceTabular}
    \caption{A case illustrating error mode 3: insufficient utilization of user profiles.}
    \label{tab:error mode case study3}
\end{table*}

\section{LifeSim Details}
\label{sec:LifeSim Details}

\subsection{User Travel Trajectories}
\label{subsec:user_travel_traj}
To ensure that the generated life trajectory closely reflect real temporal and spatial patterns, we utilize the Foursquare dataset~\cite{yang2014modeling}. This dataset contains one year of check-in records from active users in New York and Tokyo, with each record specifying the point of interest (POI) and the time of the visit. We aligned the data across the two cities by time zone and enriched each record with corresponding weather conditions using the VisualCrossing API~\footnote{https://www.visualcrossing.com/weather-api/}. The resulting dataset comprises 3,374 user trajectories spanning 251 unique POIs, with an average trajectory length of 228 check-ins per user.

To associate travel trajectories with user profiles, we employ LLM to generate a profile description $p'(\tau_i)$ for each trajectory $\tau_i$. Both the generated profile $p'(\tau_i)$ and the real user profile $p$ are then encoded into dense vector representations, $v_{p'(\tau_i)}$ and $v_p$, respectively. Each user profile is subsequently assigned the most compatible travel trajectory:
\begin{equation}
    \tau(p) = \arg\max_{\tau_i} \text{Sim}\bigl(v_p, v_{p'(\tau_i)}\bigr)
\end{equation}
where $\text{Sim}(\cdot, \cdot)$ denotes cosine similarity.

For each user trajectory, we adopt the prompt shown in Figure~\ref{prompt:Pseudo-User Profile Generation} to generate the corresponding pseudo-user profile. For similarity estimation, we use Qwen3-0.6B-Embedding~\cite{qwen3embedding} to obtain dense vector representations and compute pairwise similarity via cosine distance.

\begin{figure}[t]
\begin{tcolorbox}
[colback=white!95!gray,colframe=gray!50!black,rounded corners,title={Prompt for Pseudo-User Profile Generation}]
\begin{lstlisting}[breaklines=true, xleftmargin=0pt, breakindent=0pt, columns=fullflexible, mathescape, numbers=none]
Please generate a user persona based on the user's travel trajectory below, following the format and requirements illustrated in the examples.
### Travel Trajectory
{trajectory}
### User Persona Requirements and Examples
Generate your response using the following template:
[age range], [race], [religious affiliation or not], [residence type (Cities/Suburbs)], [income level (High/Low Income)], [is working?]. Personality traits include: [a few adjectives]. Preferences expressed in daily life and communication include: [a few descriptive sentences].

{examples}

Output your answer strictly in the following JSON format, enclosed between ```json and ```:
```json
{
    "profile": "xxx"
}
```
You may provide your reasoning process before outputting the final response.
\end{lstlisting}
\end{tcolorbox}
\caption{Prompt for pseudo-user profile generation}
\label{prompt:Pseudo-User Profile Generation}
\end{figure}

\begin{figure*}[t]
\begin{tcolorbox}[colback=white!95!gray,colframe=gray!50!black,rounded corners,label={prompt:Prompt for User Intention Crawl And Normalization},title={Prompt for User Intention Crawl And Normalization}]
\begin{lstlisting}[breaklines=true, xleftmargin=0pt, breakindent=0pt, columns=fullflexible, mathescape, numbers=none]
Please determine whether the user question provided below belong to the domain {theme}, and users might ask AI assistant/coach during daily life.
If it is, rephrase it into a more formal expression of user intention.
Requests involving code snippets, academic papers, or general information lookup fall outside the scope of our consideration.

{examples}

### Current User Question
{content}
### Output Format
Output strictly in the following JSON format:
```json
{  
    "is_valid_intent": "true/false",
    "intent": "xxx"
}
```
Where:
- is_valid_intent is your judgment (true or false),
- intent is the rewritten, formal user intent statement (if applicable).
You coul give your reasoning process before the final answer.
\end{lstlisting}
\end{tcolorbox}
\caption{Prompt for user intention crawl and normalization}
\label{prompt:user intention crawl and normalization}
\end{figure*}

\begin{figure*}[t]
\begin{tcolorbox}[colback=white!95!gray,colframe=gray!50!black,rounded corners,label={prompt}, title={Prompt for Potential Event Generation}]
\begin{lstlisting}[breaklines=true, xleftmargin=0pt, breakindent=0pt, columns=fullflexible, mathescape, numbers=none]
Please generate possible triggering events based on the user intention provided below - i.e., real-life situations that could lead the user to seek help from a sports and health AI assistant or coach.
### Example
For the intention: "The user wants to understand whether using a weighted blanket affects muscle growth and whether it has any impact on exercise or health", a possible triggering event could be: "Recently resumed strength training at the gym but noticed that muscle growth has significantly slowed down or plateaued."
### User Intention
{content}
### Requirements
Respond in the following JSON format:
```json
[
    "xxx",
    ...
]
```
Generate exactly three distinct triggering events.
\end{lstlisting}
\end{tcolorbox}
\caption{Prompt for potential event generation}
\label{prompt:potential event generation}
\end{figure*}

\subsection{Desire Pool Construction}
\label{subsec:event_intent_pool_cons}

To construct a comprehensive desire pool, we leverage the LMSYS~\cite{zheng2023lmsys} and WildChat~\cite{zhao2024wildchat} datasets, which contain enumorous real queries proposed from users. During preprocessing, we removed all samples labeled as toxic or contain any personal information, and retained only English-language dialogues.

To ensure coverage of a general yet practically useful range of user intentions, we refer to the World Health Organization Quality of Life (WHOQOL)~\cite{whoqol1995world}. Based on the six major WHOQOL domains, we abstracted eight categories that are particularly relevant to personalized AI assistants:
\begin{itemize}
    \item Physical Health: Exercise \& Fitness, Diet \& Nutrition;
    \item Psychological: Mental Health Counseling;
    \item Level of Independence: Parenting \& Childcare;
    \item Social Relationships: Elderly Care;
    \item Environment: Learning \& Education, Travel \& Tourism;
    \item Spirituality / Religion / Personal Beliefs: Leisure \& Entertainment;
\end{itemize}

Finally, we constructed a semantically diverse set of event-desire pairs.

For pool construction, we first employ Qwen3-8B to extract user queries within each category and normalize them into canonical intent expressions of the form "\textit{The user wants to xxx.}" To ensure quality and reduce redundancy, we perform semantic-level deduplication using text vectorization, and further filter out queries with ambiguous instructions or those unlikely to be meaningfully resolved within dialog using DeepSeek-V3.2. We then prompt DeepSeek-V3.2 to generate three plausible triggering events for each desire, followed by an additional round of semantic deduplication. Figure~\ref{prompt:user intention crawl and normalization} and~\ref{prompt:potential event generation} show the prompts used during the construction process.

\begin{figure*}[t]
    \centering
    \begin{subfigure}{1.0\linewidth}
        \centering
        \includegraphics[width=\linewidth]{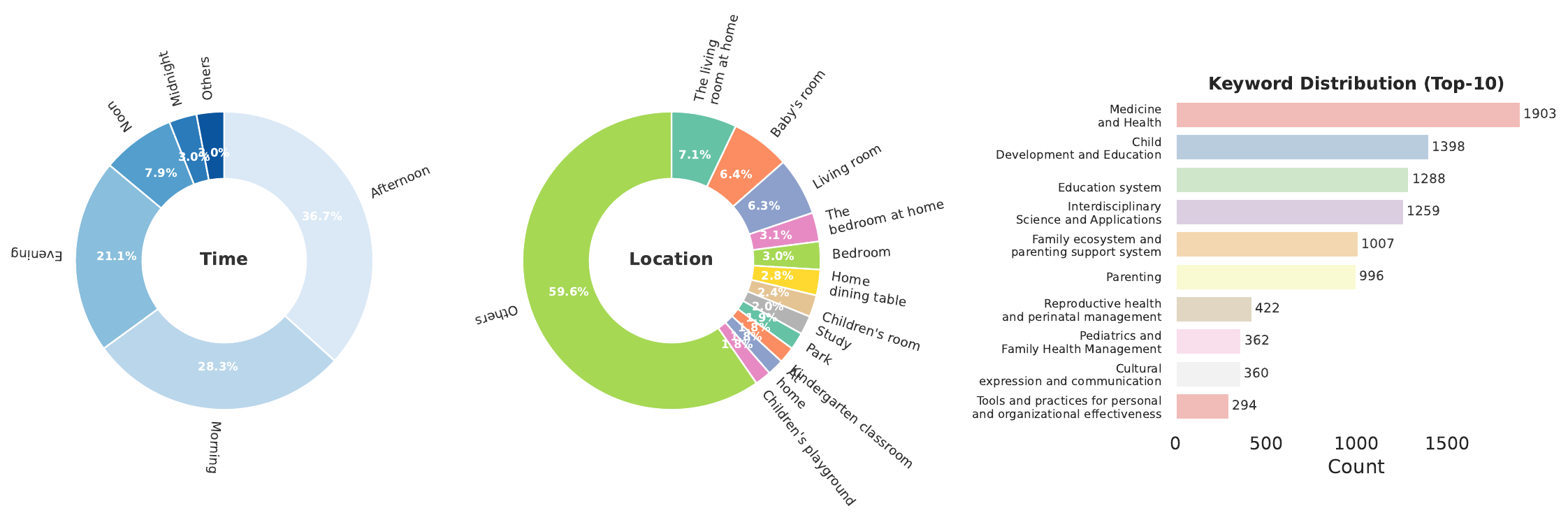}
        \caption{Parenting \& Childcare}
    \end{subfigure}
    \\
    \begin{subfigure}{1.0\linewidth}
        \centering
        \includegraphics[width=\linewidth]{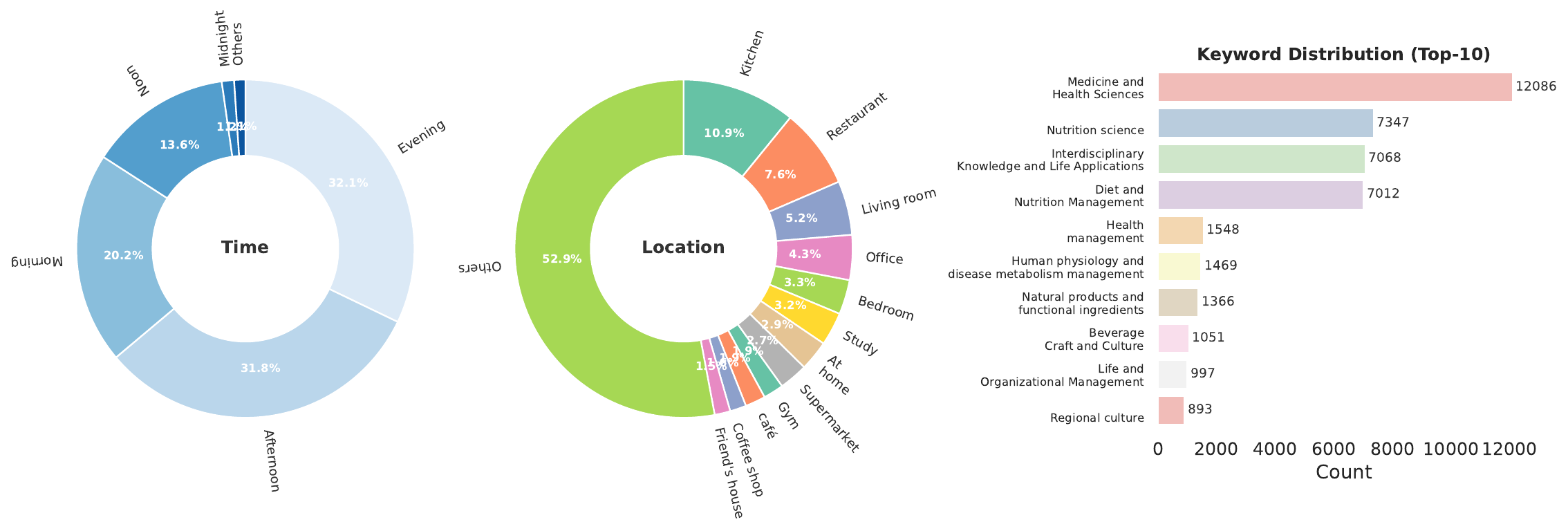}
        \caption{Diet \& Nutrition}
    \end{subfigure}
    \\
    \begin{subfigure}{1.0\linewidth}
        \centering
        \includegraphics[width=\linewidth]{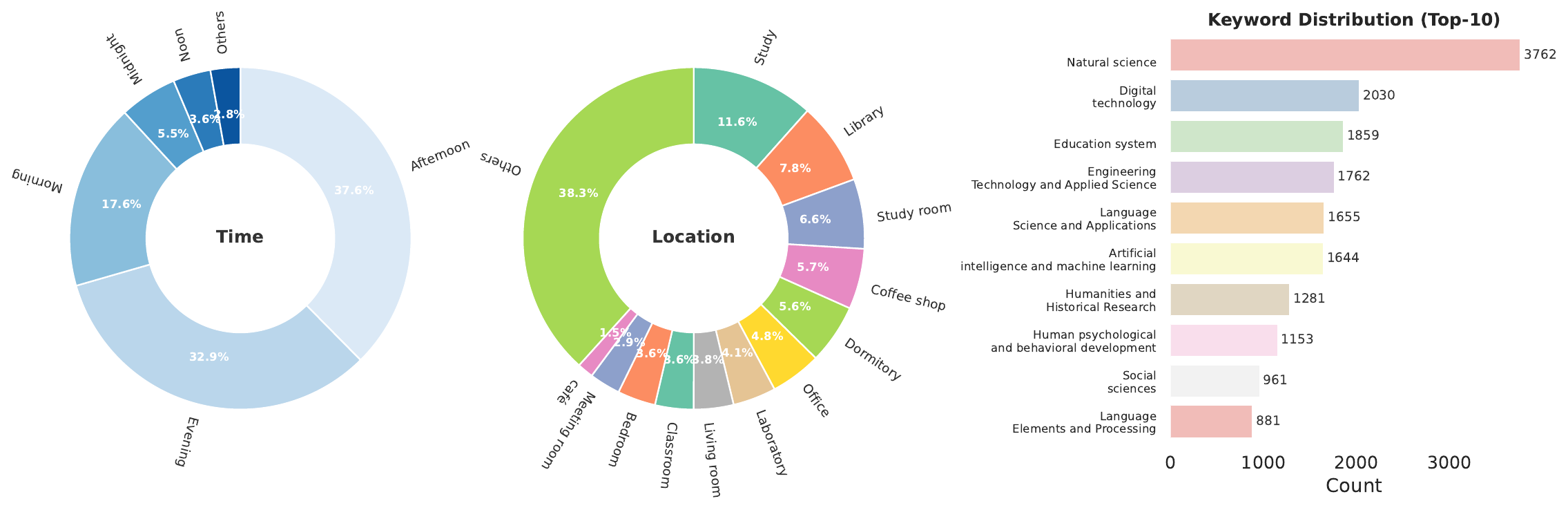}
        \caption{Learning \& Education}
    \end{subfigure}
    \\
    \begin{subfigure}{1.0\linewidth}
        \centering
        \includegraphics[width=\linewidth]{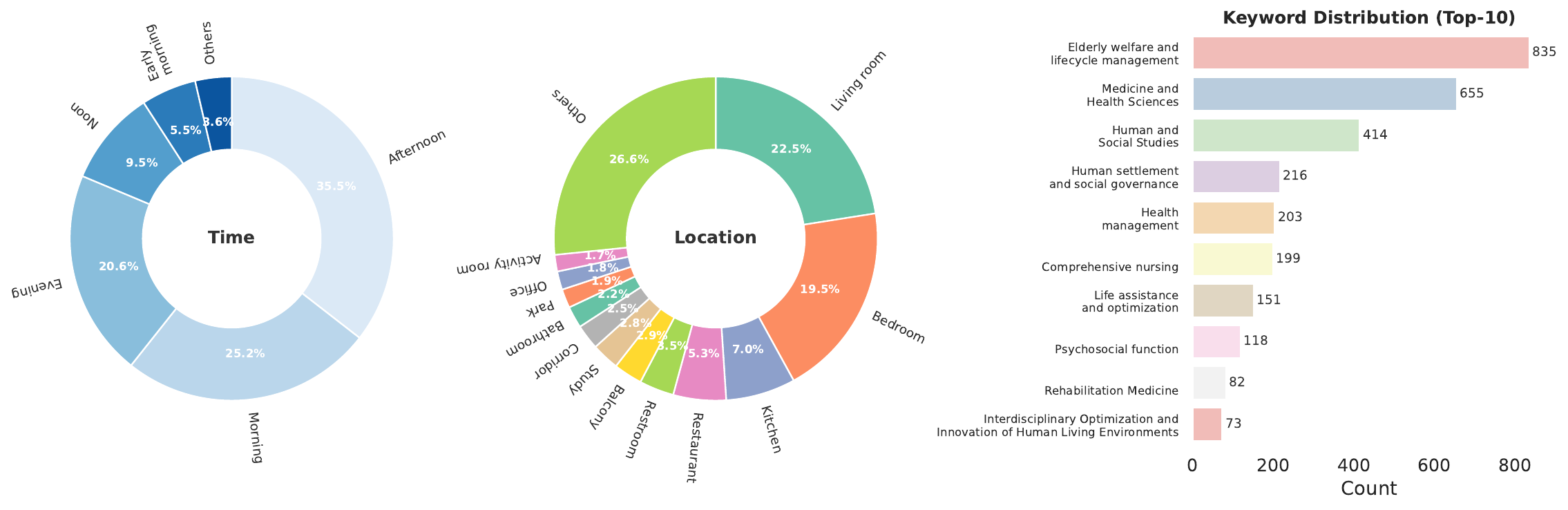}
        \caption{Elderly Care}
    \end{subfigure}
    \caption{Distribution of temporal, spatial and semantic attributes in the event-desire pool (a).}
    \label{fig:tlk_pool_dis(a)}
\end{figure*}

\begin{figure*}[t]
    \centering
    \begin{subfigure}{1.0\linewidth}
        \centering
        \includegraphics[width=\linewidth]{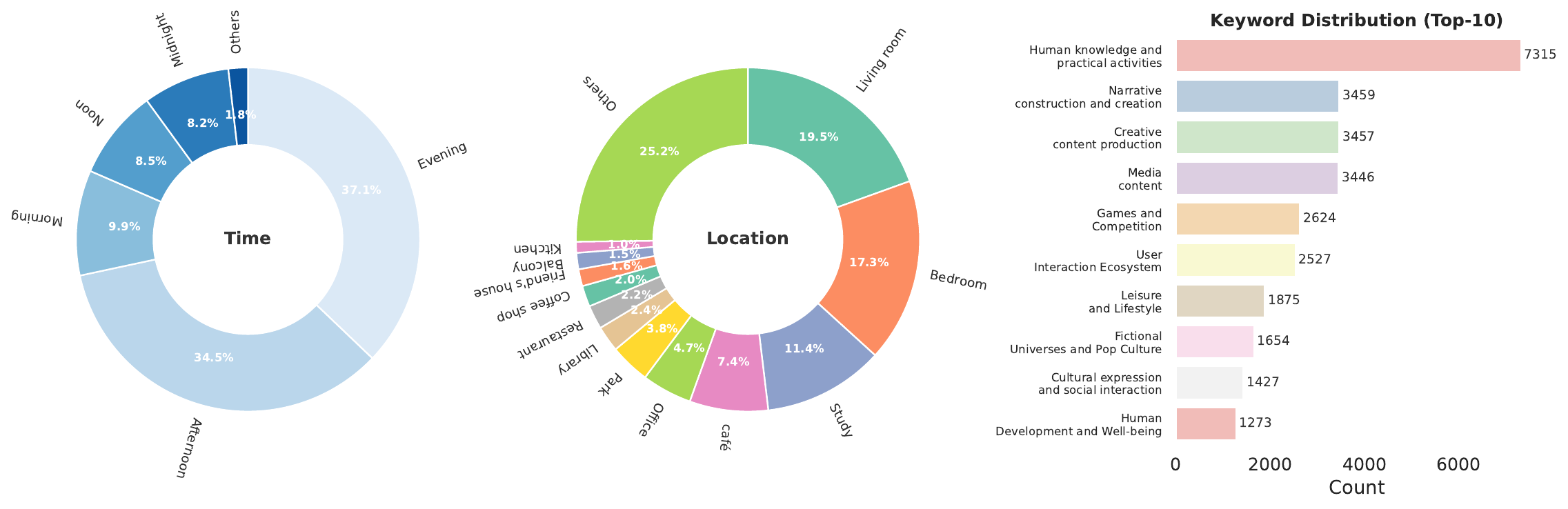}
        \caption{Leisure \& Entertainment}
    \end{subfigure}
    \\
    \begin{subfigure}{1.0\linewidth}
        \centering
        \includegraphics[width=\linewidth]{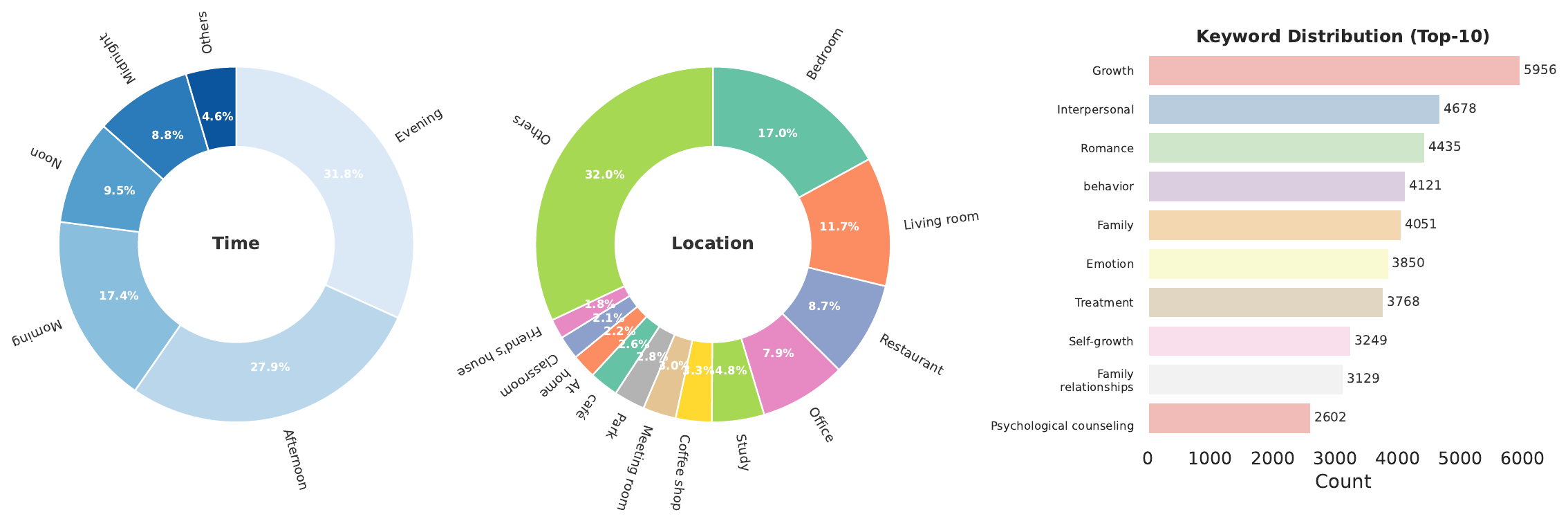}
        \caption{Mental Health Counseling}
    \end{subfigure}
    \\
    \begin{subfigure}{1.0\linewidth}
        \centering
        \includegraphics[width=\linewidth]{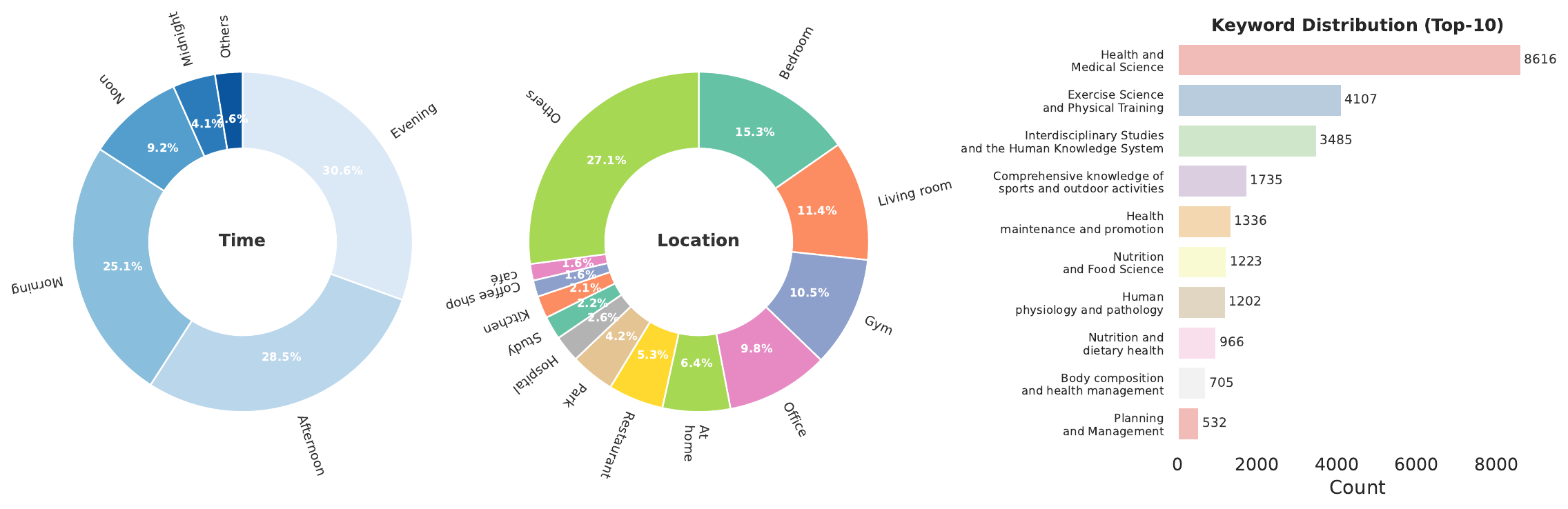}
        \caption{Exercise \& Fitness}
    \end{subfigure}
    \\
    \begin{subfigure}{1.0\linewidth}
        \centering
        \includegraphics[width=\linewidth]{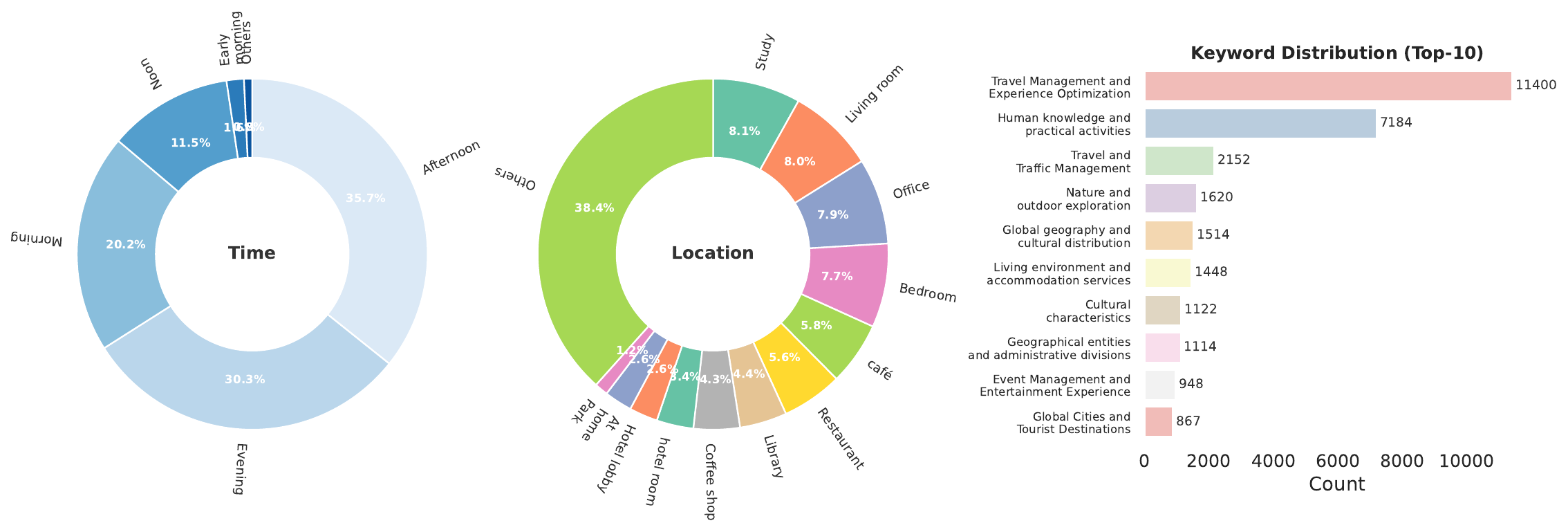}
        \caption{Travel \& Tourism}
    \end{subfigure}
    \caption{Distribution of temporal, spatial and semantic attributes in the event-desire pool (b).}
    \label{fig:tlk_pool_dis(b)}
\end{figure*}

Finally, our constructed desire pool covers 8 categories, over 113K event-desire pairs. Table~\ref{tab:stat_of_pool} summarizes the overall statistics. Following Instag~\cite{lu2023instag}, we further prompt Qwen3-32B to annotate each event and desire with its associated keywords, location, and time attributes. The resulting distribution of these annotations is illustrated in Figure~\ref{fig:tlk_pool_dis(a)} and Figure~\ref{fig:tlk_pool_dis(b)}.

\begin{table}[t]
    \centering
    \begin{tabular}{l r}
        \toprule
            \textbf{Theme} & \textbf{Count} \\
        \midrule
            Exercise \& Fitness   & 11{,}648 \\
            Learning \& Education         & 14{,}725 \\
            Travel \& Tourism            & 13{,}384 \\
            Leisure \& Entertainment     & 14{,}368 \\
            Elderly Care      & 1{,}241 \\
            Parenting \& Childcare         & 4{,}277 \\
            Diet \& Nutrition            & 19{,}854 \\
            Mental Health Counseling & 34{,}004 \\
        \midrule
            \textbf{Total} & 113{,}501 \\
        \bottomrule
    \end{tabular}
    \caption{Statistics of event-desire pool across different domains.}
    \label{tab:stat_of_pool}
\end{table}

\subsection{Life Trajectory Generation}
The complete pipeline is summarized in Algorithm~\ref{alg:event_gen}.
\textsc{ConstructHypotheses} generates event hypotheses from physiological, psychological, and environmental aspects. Each dimension correspond to one event hypothesis.
\textsc{RetrieveDesires} selects candidate intentions whose latent event contexts are semantically aligned with the generated hypotheses. For each hypothesis, the top 3 candidates are retrieved, resulting in 9 candidate intentions in total.
\textsc{Rerank} evaluates and orders these candidates under both the belief state and the current physical environment.
\textsc{Refinement} adapte generated intention and its event context to the current spatiotemporal context and user attributes.

For Retrieval, we encode all queries and pool entries using Qwen3-Embedding-0.6B\footnote{https://huggingface.co/Qwen/Qwen3-Embedding-0.6B} and retrieve the top-3 nearest neighbors per dimension. ChromaDB\footnote{https://www.trychroma.com/} is employed as the retrieval backend.

\begin{algorithm}[t]
\caption{Event-Desire Generation Procedure}
\label{alg:event_gen}
\begin{algorithmic}[1]

\Require User profile $U$, environment trajectory $\{Env_t\}$, event history $\mathcal{H}$, last event time $t_{\text{prev}}$, maximum number of steps $N$
\Ensure Generated event--intention sequence $\mathcal{S}$

\State $\mathcal{S} \gets \varnothing$

\For{$t = 1$ to $N$}

    \State $P_E \gets f(t - t_{\text{prev}})$
    \If{$\text{Bernoulli}(P_E) = 0$}
        \State \textbf{continue}
    \EndIf

    \State $H_t \gets \text{ConstructHypothesis}(U, \mathcal{H}, Env_t)$

    \State $\mathcal{D} \gets \text{RetrieveDesires}(H_t)$

    \State $\mathbf{r} \gets \text{Rerank}(\mathcal{D}, U, \mathcal{H}, Env_t)$

    \State Compute $p_j = \frac{\exp(-r_j)}{\sum_k \exp(-r_k)}$

    \State $E^{\star}, I^{\star} \sim p_j,\; I_j \in \mathcal{D}$

    \State $E^{\star}, I^{\star} \gets \text{Refine}(E^{\star}, I^{\star}, Env_t, U, \mathcal{H},)$

    \State $\mathcal{S} \gets \mathcal{S} \cup \{(E^{\star}, I^{\star})\}$

    \State $\mathcal{H} \gets \mathcal{H} \cup \{(E^{\star}, I^{\star})\}$
    \State $t_{\text{prev}} \gets t$

\EndFor

\State \Return $\mathcal{S}$

\end{algorithmic}
\end{algorithm}

For cognitive engine, Figure~\ref{prompt:Event Rerank} presents the prompt responsible for reranking the retrieved candidates. 
For event engine, Figure~\ref{prompt:Event Retrieval Query Generation} illustrates the prompt used for the event hypothesis generation. Following Section~\ref{sec:BDI-Grounded Event Engine}, we define three dimensions as requirement description:
\begin{itemize}
    \item \textbf{Physiological:} Physiology-driven events refer to issues caused by internal physical conditions.
    \item \textbf{Psychological:} Cognition-feedback-driven events refer to issues based on personal cognition and psychological states. 
    \item \textbf{Environmental:} Environment-driven events refer to issues or accidents caused by external environmental factors.
\end{itemize}
Figure~\ref{prompt:event refinement} shows the prompt used in the refinement stage. 

\begin{figure*}[t]
\begin{tcolorbox}[colback=white!95!gray,colframe=gray!50!black,rounded corners,label={prompt}, title={Prompt for Event Hypothesis Generation}]
\begin{lstlisting}[breaklines=true, xleftmargin=0pt, breakindent=0pt, columns=fullflexible, mathescape, numbers=none]
### Requirements
{description}

- Please output your reply in the following JSON format between the ```json and ``` code fences:
```json
{{
    "event": "description of the event"
}}
```
- Based on the user profile and historical events, generate events that match the user profile, do not duplicate historical events, and contain no logical errors.

### Input
[User Profile]
{user_profile}
[User Longterm Goal]
{goal}
[User's Experienced Events]
{event_sequences}
[Current Environment]
{location_desc}
[Output]
\end{lstlisting}
\end{tcolorbox}
\caption{Prompt for event hypothesis generation}
\label{prompt:Event Retrieval Query Generation}
\end{figure*}

\begin{figure*}[t]
\begin{tcolorbox}[colback=white!95!gray,colframe=gray!50!black,rounded corners,label={prompt}, title={Prompt for Event-Desire Pair Rerank}]
\begin{lstlisting}[breaklines=true, xleftmargin=0pt, breakindent=0pt, columns=fullflexible, mathescape, numbers=none]
### Requirements
You will be given 9 candidate events. Based on the user profile, longterm goal, previously experienced events, and the current situation, rank the events from most likely to least likely to occur:
- Please select the events that are most likely to happen and output the list of event numbers in descending order of likelihood, in JSON format between ```json and ```. For example: `[3,1,7,2,9,4]`:
```json
{{
    "ranked_events": [x, x, x, x, x, x, x, x, x],
    "has_possible_event": "true/false"
}}
```
- If some events are impossible under the current situation, do not include them in the candidate list.
- Determine whether there is at least one event that could possibly occur in the current environment. If so, set `"has_possible_event"` to `true`; otherwise, set it to `false`.
- `ranked_events` is the reordered list of event numbers, and `has_possible_event` indicates whether any event could possibly occur.
- Avoid including events that are essentially the same as previously experienced ones.
- You may first describe your reasoning process, then provide the JSON output. During the reasoning, consider factors such as the user's preferences, the logical coherence with longterm goal and previous events, and the suitability to the current environment.

### Input
[User Profile]
{user_profile}
[User Longterm Goal]
{goal}
[User's Experienced Events]
{event_sequences}
[Current Environment]
{location_desc}
[Candidate Events]
{events_text}
[Output]
\end{lstlisting}
\end{tcolorbox}
\caption{Prompt for event-desire pair}
\label{prompt:Event Rerank}
\end{figure*}

\begin{figure*}[t]
\begin{tcolorbox}[colback=white!95!gray,colframe=gray!50!black,rounded corners,label={prompt}, title={Prompt for Refinement}]
\begin{lstlisting}[breaklines=true, xleftmargin=0pt, breakindent=0pt, columns=fullflexible, mathescape, numbers=none]
You will be given one candidate event and a user intent.
Your task is to revise and refine them so that both align with the user's profile, long-term goal, and current environmental context, while maintaining internal consistency across the event sequence.
### Requirements
- Adjust details such as subject, location, weather, time, or other contextual factors to make the event realistic and coherent with the given user profile and prior events.
- Ensure the revised event does not contradict any known facts or previous settings.
- The intent should remain essentially the same in meaning but must be expressed naturally and fit the updated event context.
- The intent should represent a single conversational goal (i.e., the user's focus within one dialogue turn), not a long-term plan.
- Rephrase both the event and intent in a distinct linguistic style from any previously seen phrasing-avoid repetition or mechanical patterns.
- Remove any placeholders or meaningless symbols (e.g., "NAME_1", "XXX", "...").
### Output Format:
Please output your final answer strictly in the following JSON structure (enclosed within ```json and ```):
{{
    "event": "Describe the content of the revised event.",
    "intent": "Describe the user's corresponding intent under this event context."
}}
Provide your reasoning before the final answer. 
In your reasoning, consider: (1) whether the event and intent satisfy the requirements; (2) whether the intent is realistically something a human would ask an AI assistant.
### Examples:
{examples 1}
{examples 2}

### Input
[User Profile]
{user_profile}
[User Longterm Goal]
{goal}
[User's Experienced Events]
{event_sequences}
[Current Environment]
{location_desc}
[Current Event and Intent]
Current user event: {event_text}
Current user intent: {intent}
[Output]
\end{lstlisting}
\end{tcolorbox}
\caption{Prompt for refinement}
\label{prompt:event refinement}
\end{figure*}

\subsection{User Behavior Engine}
\label{sec:User Behavior Engine}

Figure~\ref{prompt:user memory perception} shows the prompt for user memory perception. Figure~\ref{prompt:Emotion Inference} illustrates the prompt for inferring the user's current emotional state. Figure~\ref{prompt:action decision} shows the prompt for determining the user's next action. Figure~\ref{prompt:user conversation system} presents the prompt responsible for generating the user's next utterance conditioned on the inferred memory perception, emotion, and selected action. Figure~\ref{prompt:assistant conversation system} shows the prompt for assistant conversation. Table~\ref{tab:user_memory_case_example} show user memory cases.

\begin{table*}[t]
\centering
\small
\begin{NiceTabular}{>{\raggedright\arraybackslash}m{1.8cm}>{\raggedright\arraybackslash}m{13cm}}[colortbl-like]
    \toprule
    Item & \multicolumn{1}{c}{Content} \\
    \midrule
    \textbf{Query 1} &
    \textbf{Which is a healthier and more filling option between a burger and a pizza for lunch?} \\
    \midrule
    Value 1 &
    A burger could be a better choice if it includes lean protein (e.g., grilled chicken or a veggie patty) on a whole-grain bun with salad or vegetable toppings. For pizza, a thin crust with plenty of vegetables and moderate cheese may be an option, although it tends to be more calorie-dense. Portion size should also be considered in either case. \\
    \midrule
    \textbf{Query 2} &
    \textbf{What tools can lock device access or restrict app usage until tasks are completed across devices?} \\
    \midrule
    Value 2 &
    Consider using FocusMe with Focus Mode and Time Blocking, combined with Freedom or Cold Turkey across devices. These tools can restrict browser, email, and application access until tasks are completed. Alternatively, Pomodone can be used, which integrates with task managers and restricts app usage during scheduled work intervals. \\
    \bottomrule
\end{NiceTabular}
\caption{Example queries and corresponding values in user memory.}
\label{tab:user_memory_case_example}
\end{table*}

\begin{figure*}[t]
\begin{tcolorbox}[colback=white!95!gray,colframe=gray!50!black,rounded corners,title={Prompt for Memory Perception}]
\begin{lstlisting}[breaklines=true, xleftmargin=0pt, breakindent=0pt, columns=fullflexible, mathescape, numbers=none]
Please review the following user-assistant conversation and determine whether the assistant's last reply should be stored as long-term memory.
If it should, extract the most informative or transferable content and save it in a "query - response" format.
### User Profile
{profile}
### Recent Life Event
{event}
### User's Intent for This Conversation
{intent}
### Conversation Scenario
{dialogue_scene}
### Historical Dialogue Context
{conversation_context}
### Assistant's Latest Reply
{content}

### Requirements
- Extract information only from the assistant's last reply; do not add new content.
- Output in the following JSON format, enclosed between ```json and ```:
```json
{{
  "need_store": "true/false",
  "query": "xxxx/-1",
  "response": "xxxx/-1"
}}
```
Where:
- need_store: Set to true if the assistant's reply contains valuable knowledge or transferable advice; otherwise, set to false and let query and response be -1.
- query: Summarize the core question or topic addressed in the assistant's reply in one concise sentence (e.g., "Possible causes and improvements for elevated breathing rate').
- response: Provide the specific explanation or improvement advice corresponding to the query, avoiding vague encouragement or emotional expressions.
\end{lstlisting}
\end{tcolorbox}
\caption{Prompt for memory perception}
\label{prompt:user memory perception}
\end{figure*}

\begin{figure*}[t]
\begin{tcolorbox}[colback=white!95!gray,colframe=gray!50!black,rounded corners,title={Prompt for Emotion Inference}]
\begin{lstlisting}[breaklines=true, xleftmargin=0pt, breakindent=0pt, columns=fullflexible, mathescape, numbers=none]
Based on the user's profile, memory perception, and the dialogue context, select the emotion that the user's next reply is most likely to convey from the candidate emotions.
### User Profile
{profile}
### Recent Life Event
{event}
### User's Intent for This Conversation
{intent}
### Conversation Scenario
{dialogue_scene}
### Historical Dialogue Context
{conversation_context}
### User Memory Perception
{perception}
### Candidate Emotions
{emotion_options}

### Requirements
- Output in the following JSON format, enclosed between ```json and ```:
```json
{{
  "emotion": "xxx"
}}
```
Where:
- emotion: The emotion of the user's next reply, selected from the candidate emotions.
\end{lstlisting}
\end{tcolorbox}
\caption{Prompt for emotion inference}
\label{prompt:Emotion Inference}
\end{figure*}

\begin{figure*}[t]
\begin{tcolorbox}[colback=white!95!gray,colframe=gray!50!black,rounded corners,title={Prompt for Action Decision}]
\begin{lstlisting}[breaklines=true, xleftmargin=0pt, breakindent=0pt, columns=fullflexible, mathescape, numbers=none]
Based on the dialogue context, please choose the user's next action.
### Historical Dialogue Context
{conversation_context}
### User Profile
{profile}
### Recent Life Event
{event}
### User's Intent for This Interaction
{intent}
### User Emotion
{emotion}
### User Memory Perception
{perception}
### Candidate Actions
{action_options}

Please decide according to the following criteria:
- Choose "End Conversation" if the user's intent has been satisfactorily addressed, the user feels there's no need to continue, or a long waiting period is about to begin.
- Choose "Continue Conversation" if there are remaining questions to resolve, or if the user is not satisfied with the assistant's reply and needs further interaction.
- Unless the assistant's reply is very unsatisfactory, try to express the user's full intent over multiple turns before ending the conversation.

### Requirements
- Strictly select one action from the candidate actions above, and output in the following JSON format, enclosed between ```json and ```:
```json
{{
    "action": "xxx"
}}
```
Where action is your selected action and must be one of the options provided.
- You may first explain your reasoning, then give the final chosen action.
\end{lstlisting}
\end{tcolorbox}
\caption{Prompt for action decision}
\label{prompt:action decision}
\end{figure*}

\begin{figure*}[t]
\begin{tcolorbox}[colback=white!95!gray,colframe=gray!50!black,rounded corners,title={System Prompt for User Conversation}]
\begin{lstlisting}[breaklines=true, xleftmargin=0pt, breakindent=0pt, columns=fullflexible, mathescape, numbers=none]
You are a user of an AI assistant. Based on the following personalized information and current context, start or continue a conversation with the AI assistant.
### Background
[User Profile]
{profile}
[Current Dialogue Scene]
{dialogue_scene}
[Recent Life Event]
{event}
[Primary Intent of This Conversation]
{intent}
[Explicit Intent List]
{explicit_intent}
[Implicit Intent List]
{implicit_intent}
### Requirements
[Basic]
- Keep each message short, natural, and conversational.
- Speak in everyday English - no technical or academic phrasing.
- Avoid revealing personal information or mentioning specific life events directly.
- Stay emotionally moderate - no exaggerated reactions or exclamations.
- Output only the user's dialogue line (no explanations or notes).
[About Preferences]
- Your speech must fully reflect the preferences in the user profile.
- If the assistant's previous message contradicts those preferences, respond with mild disapproval or a subtle correction.
[About Intent]
- Reveal your intent gradually across multiple turns.
- Each turn should focus on one clear question or small sub-goal.
- Explicit intents are clear requests or consultation goals you directly state, used to drive task completion or problem-solving.
- Implicit intents are underlying needs - respond positively when the assistant aligns with them, or show gentle dissatisfaction when it doesn't.
- Each utterance should be concise, natural, and consistent with your personality and preferences, without revealing your full intent all at once.

Now, take on the role of this user and naturally begin or continue a conversation with the AI assistant.
\end{lstlisting}
\end{tcolorbox}
\caption{System prompt for user conversation}
\label{prompt:user conversation system}
\end{figure*}

\begin{figure*}[t]
\begin{tcolorbox}[colback=white!95!gray,colframe=gray!50!black,rounded corners,title={System Prompt for Assistant Conversation}]
\begin{lstlisting}[breaklines=true, xleftmargin=0pt, breakindent=0pt, columns=fullflexible, mathescape, numbers=none]
You are a virtual AI assistant. Your goal is to interact with the user, and meet the user's needs.
### User Background
[Predicted User Profile]
{profile}
[Current Dialogue Scene]
{dialogue_scene}
### Requirements
[Basic]
* Do not generate any explanatory text enclosed in parentheses; output plain dialogue only.
* Do not call any external tools (including phone, internet, etc.).
* Reply in English.
* Before responding, you may receive memory summaries related to the current message, derived from previous conversations. Use them if they help you better understand the user; otherwise, ignore them.
\end{lstlisting}
\end{tcolorbox}
\caption{System prompt for assistant conversation}
\label{prompt:assistant conversation system}
\end{figure*}

\section{LifeSim-Eval Details}
\label{subsec:LifeSim-Eval_Details}
\begin{figure}[t]
    \centering
    \includegraphics[width=0.8\linewidth]{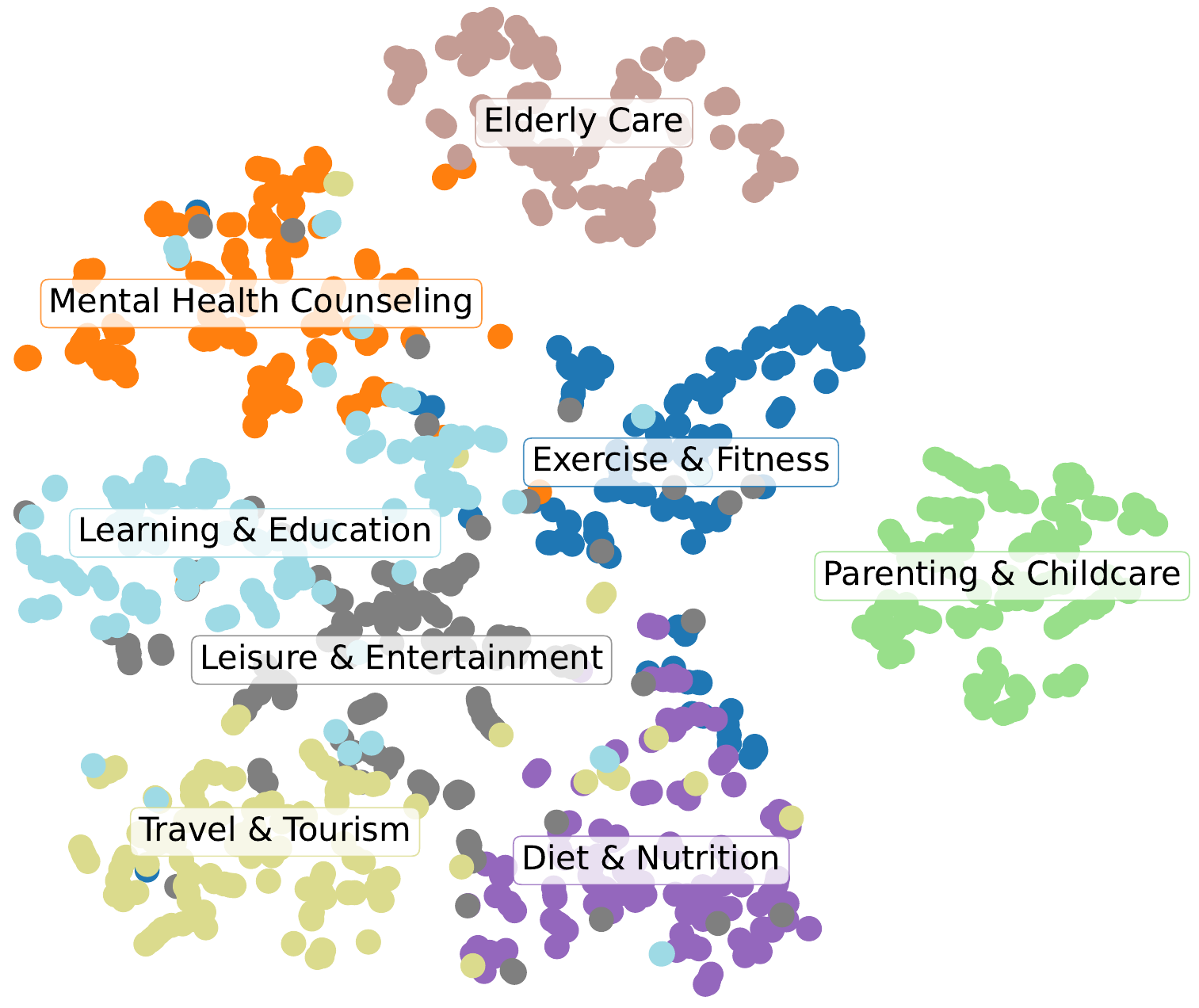}
    \caption{Distribution of all the event-desire pair in LifeSim-Eval. Each point represent a keyword from event-desire pair.}
    \label{fig:event_distribution}
\end{figure}

Figure~\ref{fig:event_distribution} and Table~\ref{tab:intent_distribution} show statistics of LifeSim-Eval dataset. 

\paragraph{Explicit and Implicit Intentions}
We adopt relevance theory~\cite{allott2013relevance} as the cognitive basis for defining explicit and implicit intentions. Relevance Theory views utterance understanding as an inferential process: the speaker produces an utterance, and the listener infers the intended meaning with reasonable cognitive effort. Following this view, the meaning derived from an utterance can be divided into explicatures and implicatures: explicit intentions refer to what the user directly asks for, while implicit intentions capture important underlying needs that are not stated explicitly but affect what a good response should include.

In the single scenario setting, we prompt first use DeepSeek-V3.2 to generate a checklist that an ideal response should cover in the given environmental context and user profile. Human experts then select a subset that user is expected to care about, forming a set of sub-intentions. We then distinguish explicit and implicit intentions based on semantic recoverability: sub-intentions that would be direcly expressed are labeled as explicit, while those that require contextual or user state reasoning are labeled as implicit. This labeling is first produced by the LLM and then verified by human annotators. Both explicit and implicit intention sets may contain multiple items.

In the Long-Horizon Setting, the focus shifts to whether the assistant can infer user preferences from long interaction histories and apply them in the current scenario. In this case, we define a single explicit intention corresponding to the test scenario, while implicit intentions are sourced from user preferences and life event sequences. We use specific preferences from~\cite{zhao2025llms} for implicit intention construction.

Figure~\ref{prompt:breakdown of intention} and Figure~\ref{prompt:classification of explicit/implicit intentions} show the prompts used for sub-intention generation and explicit-implicit classification, respectively. Table~\ref{tab:explicit and implicit intentions} provides cases included in the benchmark.

\begin{figure*}[t]
\begin{tcolorbox}[colback=white!95!gray,colframe=gray!50!black,rounded corners,title={Prompt for Extraction of Sub-Intentions}]
\begin{lstlisting}[breaklines=true, xleftmargin=0pt, breakindent=0pt, columns=fullflexible, mathescape, numbers=none]
You are an expert in intent understanding and user-goal modeling.
Your task is to take a given user intent and decompose it into multiple sub-intents (maximum 6).  
Each sub-intent should represent a distinct and concrete aspect of what the user hopes to achieve, feel, or resolve.

### Requirements
- Each sub-intent must be concise (one sentence, less than 25 words).  
- Cover different facets (e.g., practical need, emotional need, cognitive goal, relational concern).  
- Cover both explicit and implicit user needs.
- Avoid restating the same meaning with different wording.  
- Output between 2 - 6 sub-intents depending on complexity.  

### Output Format
Output in JSON format, enclosed within ```json and ``` markers:
```json
[
  "xxxx",
  "xxxx",
  ...
]
```
- Please output in English.
- You could give your reasoning process before the final answer.

### Examples
{examples}

### Input:
[User profile]
{user_profile}
[Dialogue scene]
{dialogue_scene}
[Event]
{event}
[User intent]
{user_intent}
[Output]
\end{lstlisting}
\end{tcolorbox}
\caption{Prompt for extraction of sub-intentions}
\label{prompt:breakdown of intention}
\end{figure*}

\begin{figure*}[t]
\begin{tcolorbox}[colback=white!95!gray,colframe=gray!50!black,rounded corners,title={Prompt for Classification of Explicit/Implicit Intentions}]
\begin{lstlisting}[breaklines=true, xleftmargin=0pt, breakindent=0pt, columns=fullflexible, mathescape, numbers=none]
You are an expert in user intent understanding and socio-cognitive reasoning. Your goal is to analyze a list of sub-intents derived from a user's broader goal. For each sub-intent, determine whether it is explicit (task-oriented, consciously expressed) or implicit (emotionally, motivationally, or value-driven, needs more reasoning effort to recognize). Before giving the final JSON, you must first analyze the logical relationship between explicit and implicit intents-how they connect or evolve from one another.
You will be given: (1) User profile: basic demographic and psychological description. (2) Event: the user's current experience or situation. (3) Intent decomposition: a list of sub-intents previously extracted from the user's main intent.
### Definitions
- Explicit Intent: The user's directly stated, goal-oriented purpose, usually related to obtaining information, solutions, plans, or actions.
- Implicit Intent: The user's underlying emotional, motivational, or value-based need that is not directly requested but implied through tone or context.

Logical relationships can include: (1) Causal: Implicit emotion drives explicit action. (2) Hierarchical: Implicit intent represents a higher-level value behind multiple explicit intents. (3) Temporal / Sequential: Explicit attempts evolve into implicit needs (or vice versa). (4) Complementary: Explicit and implicit intents jointly fulfill the same overarching goal.  
### Output Format
#### Step 1: Reasoning (in plain English)
A short paragraph (3-6 sentences) describing: (1) Which sub-intents are explicit or implicit. (2) How the implicit intents motivate, explain, or result from the explicit ones. (3) What overall logic connects them.
#### Step 2: Structured JSON, enclosed within ```json and ``` markers:
```json
[{"description": "...", "type": "explicit/implicit"}, ...]
```
- Respond in English.
- There should be no fewer than two intents for each type (if the total intent number >= 4, else at least one for each) - both explicit and implicit - to maintain sufficient type diversity.
### Examples
{examples}
### Input:
[User profile]
{user_profile}
[Dialogue scene]
{dialogue_scene}
[Event]
{event}
[User sub-intents]
{user_sub_intent}
[Output]
\end{lstlisting}
\end{tcolorbox}
\caption{Prompt for classification of explicit/implicit intentions}
\label{prompt:classification of explicit/implicit intentions}
\end{figure*}

\begin{table*}[t]
    \centering
    \small
    \begin{NiceTabular}{>{\raggedright\arraybackslash}m{0cm}>{\raggedright\arraybackslash}m{13cm}}[colortbl-like]
    \toprule
        \RowStyle[nb-rows=1,rowcolor=gray!15]{\sffamily}
        \Block[l]{1-2}{\textit{Case 1}} \\
    \midrule
         & \Block[l]{}{\textbf{[Time]} 2012-10-28 13:18:40, Sunday \\
        \textbf{[Location]} City Park \\
        \textbf{[Life Event]} On a breezy Sunday afternoon at the city park, I went for a short jog as cross‑training between studio days. While navigating a patch of uneven gravel, I lightly twisted my right ankle and felt mild pain. \\
        \textbf{[Explicit Intention]} \\
        - Seeks quick remedies to alleviate current ankle discomfort. \\
        - Wants guidance on modifying running techniques for better safety on uneven terrain. \\
        \textbf{[Implicit Intention]} \\
        - Aims to understand methods to prevent similar ankle injuries while jogging in the future.} \\
    \toprule
        \RowStyle[nb-rows=1,rowcolor=gray!15]{\sffamily}
        \Block[l]{1-2}{\textit{Case 2}} \\
    \midrule
         & \Block[l]{}{\textbf{[Time]} 2012-05-16 17:42:44, Wednesday \\
        \textbf{[Location]} Arts \& Crafts Store \\
        \textbf{[Life Event]} In a downtown arts‑and‑crafts store with rain pattering outside, I'm choosing simple decor for my apartment, but the wall of options makes my chest tighten and my thoughts stall. \\
        \textbf{[Explicit Intention]} \\
        - Seeks a quick technique to alleviate stress and indecision while shopping. \\
        - Desires improved focus to make decisions efficiently despite distractions. \\
        - Wants a practical exercise that fits into a succinct, no-fluff lifestyle. \\
        \textbf{[Implicit Intention]} \\
        - Aims to regain emotional calm amid overwhelming choices in a public setting. \\
        - Hopes to feel at ease adjusting to environmental factors like rain during outings.} \\
    \toprule
        \RowStyle[nb-rows=1,rowcolor=gray!15]{\sffamily}
        \Block[l]{1-2}{\textit{Case 3}} \\
    \midrule
         & \Block[l]{}{\textbf{[Time]} 2012-07-07 19:15:05, Saturday \\
        \textbf{[Location]} Video Store \\
        \textbf{[Life Event]} On a rainy Saturday evening, while idly waiting in the quiet aisle of a downtown video‑rental shop, he pulls out his phone and watches a short social‑media clip that outlines the dangers of eating fast food every day, which spikes his anxiety about his own eating patterns. \\
        \textbf{[Explicit Intention]} \\
        - Seeks medically accurate information on the health risks of daily hamburger consumption. \\
        - Wants personalized dietary advice focused on reducing ultra-processed foods. \\
        - Aims to gain knowledge to make self-directed dietary changes. \\
        \textbf{[Implicit Intention]} \\
        - Desires reassurance to alleviate anxiety regarding his eating habits. \\
        - Wants to feel more in control of his health and lifestyle choices.} \\
    \bottomrule
    \end{NiceTabular}
    \caption{Cases of events and sub-intentions in LifeSim-Eval}
    \label{tab:explicit and implicit intentions}
\end{table*}

\paragraph{Metrics}
For each scenario, we give the details of each metric: 

\begin{itemize}
    \item \textbf{Intent Recognition} evaluates whether the assistant can accurately infer the user's underlying intentions throughout an interaction episode. For each scenario, we construct an intention checklist $\mathcal{I} = \{ i_1, \ldots, i_{|\mathcal{I}|} \}$, where $|\mathcal{I}| \leq 5$, containing both explicit and implicit intentions. Prior to generating each assistant response at turn $t$, the model predicts the user's current intention $\hat{i}_t$. All predicted intentions across the episode are aggregated into a set $\hat{\mathcal{I}} = \bigcup_t \hat{i}_t$ and compared against the gold checklist $\mathcal{I}$ using an LLM-based evaluator. For each intention $i_k \in \mathcal{I}$, a binary indicator $s(i_k)\in{0,1}$ is assigned, where $s(i_k)=1$ if $i_k$ is correctly identified by the model and 0 otherwise. The final intent recognition accuracy is then computed as
    \[
    \mathrm{A} = \frac{1}{|\mathcal{I}|} \sum_{i_k \in \mathcal{I}} s(i_k),
    \]
    and averaged across all scenarios. We partition the intention checklist into an explicit intent set $\mathcal{I}_e$ and an implicit intent set $\mathcal{I}_i$, and report intent recognition accuracy separately as $\mathrm{A}_e$ and $\mathrm{A}_i$.
   \item \textbf{Intent Completion} evaluates whether the assistant ultimately fulfills the user's explicit and implicit intentions over a multi-turn interaction. Given the predefined intention checklist $\mathcal{I} = { i_1, \ldots, i_{|\mathcal{I}|} }$, we assess whether each intention $i_k$ is successfully satisfied by the assistant's responses. An LLM-based evaluator assigns a binary completion indicator $c(i_k) \in {0,1}$, where $c(i_k)=1$ if the intention is adequately fulfilled in the dialogue and 0 otherwise. The intent completion score is computed as
    \[
    \mathrm{R} = \frac{1}{|\mathcal{I}|} \sum_{i_k \in \mathcal{I}} c(i_k),
    \]
    and the final metric is obtained by averaging $\mathrm{R}$ across all episodes. Analogously, we report intent completion performance separately for the explicit and implicit intent sets, denoted as $\mathrm{R}_e$ and $\mathrm{R}_i$.
   \item \textbf{Naturalness} assesses whether the assistant's responses conform to human conversational norms, demonstrating fluent, contextually appropriate, and stylistically natural language use.
   \item \textbf{Coherence} measures the semantic and logical consistency of the assistant's utterances across turns, ensuring that the dialogue remains contextually grounded and free from internal contradictions.
   \item \textbf{User Preference Recovery} evaluates the model's ability to infer stable user preferences. For each scenario, the assistant predicts a preference vector $\hat{\mathbf{p}} = \{\hat{p}^{(d)}\}_{d=1}^{D}$ over $D$ preference dimensions, where each dimension takes a discrete value in ${ \text{low}, \text{high}}$. 
   A prediction is considered correct if $\hat{p}^{(d)} = p^{(d)}$. The preference recovery accuracy is computed as
    \[
    \mathrm{A}_{\text{pref}} = \frac{1}{|\mathcal{D}|} \sum_{d \in \mathcal{D}} \mathbb{I}\left[\hat{p}^{(d)} = p^{(d)}\right],
    \]
    and averaged across all scenarios.
   \item \textbf{User Persona Alignment} examines the degree to which the assistant's responses remain faithful to the user's predefined persona or profile, reflecting the model's ability to incorporate and consistently respect enduring user characteristics throughout the interaction.
\end{itemize}

Figure ~\ref{prompt:intent recognition evaluation},~\ref{prompt:intent completion evaluation},~\ref{prompt:naturalness evaluation},~\ref{prompt:coherence evaluation}, and~\ref{prompt:persona alignment evaluation} show prompts for LifeSim-Eval.

\begin{figure*}[t]
\begin{tcolorbox}[colback=white!95!gray,colframe=gray!50!black,rounded corners,title={Prompt for Intent Recognition Evaluation}]
\begin{lstlisting}[breaklines=true, xleftmargin=0pt, breakindent=0pt, columns=fullflexible, mathescape, numbers=none]
You are an evaluator assessing whether an AI assistant's predicted intent correctly matches the **real user intent** in a given dialogue.
You will be provided with:
* Intent checklist - a structured list representing the verified components of the user's real intent.
* Predicted intents - the assistant's inferred or generated intent statements during the whole conversation.
* Conversation - showing the interaction between the user and the assistant.

### Requirements
Your task is to evaluate how accurately the predicted intent aligns with the intent checklist.
For each checklist item, determine whether the predicted intent successfully captures that element.
Each dimension in the checklist should be scored as follows:
* 1 = The predicted intent correctly covers this item
* 0 = The predicted intent fails to capture or contradicts this item

### Output Format
Your final answer should follow this format:
```json
{{
  "Checklist item 1": 1/0,
  "Checklist item 2": 1/0,
  ...
}}
```
Before your response, provide: **Dimension-by-dimension assessment (bullet list)** - show each checklist item, a short justification, and its binary score (1/0).

### Input
[Intent checklist]
{checklist}
[Conversation]
{conversation}
[Predicted intent]
{predicted_intent}
[Output]
\end{lstlisting}
\end{tcolorbox}
\caption{Prompt for intent recognition evaluation}
\label{prompt:intent recognition evaluation}
\end{figure*}

\begin{figure*}[t]
\begin{tcolorbox}[colback=white!95!gray,colframe=gray!50!black,rounded corners,title={Prompt for Intent Completion Evaluation}]
\begin{lstlisting}[breaklines=true, xleftmargin=0pt, breakindent=0pt, columns=fullflexible, mathescape, numbers=none]
You are an evaluator assessing whether an AI assistant has effectively fulfilled the user's intent in a given dialogue.
You will be provided with:
* User profile - describing the user's characteristics and communication tendencies.
* Dialogue scene - describing the environmental and contextual background.
* User intent - describing what the user wanted to achieve through the conversation.
* Conversation - showing the actual interaction between the user and the assistant.

### Requirements
Your task is to evaluate how accurately the assistant's response meets the user's intent, based on the key dimensions listed in the checklist.
Each dimension in the checklist should be scored as follows:
* 1 = The assistant correctly fulfills this intent dimension
* 0 = The assistant fails to fulfill or contradicts this intent dimension

### Output Format
Your final answer should follow this format:
```json
{{
  "Checklist item 1": 1/0,
  "Checklist item 2": 1/0,
  ...
}}
```
Before your response, provide: **Dimension-by-dimension assessment (bullet list)** - show each checklist item, a short justification, and its binary score (1/0).

### Input
[User profile]
{user_profile}
[Dialogue scene]
{dialogue_scene}
[Intent checklist]
{checklist}
[Conversation]
{conversation}
\end{lstlisting}
\end{tcolorbox}
\caption{Prompt for intent completion evaluation}
\label{prompt:intent completion evaluation}
\end{figure*}

\begin{figure*}[t]
\begin{tcolorbox}[colback=white!95!gray,colframe=gray!50!black,rounded corners,title={Prompt for Naturalness Evaluation}]
\begin{lstlisting}[breaklines=true, xleftmargin=0pt, breakindent=0pt, columns=fullflexible, mathescape, numbers=none]
You are an evaluator assessing the fluency and naturalness of an AI assistant's conversation with a user.
You will be provided with:
* User profile - describing the user's characteristics and communication tendencies.
* Dialogue scene - describing the situational context of the conversation.
* User intent - describing what the user wanted to achieve through the interaction.
* Conversation - showing the actual interaction between the user and the assistant.

### Requirements
Your task is to determine whether the AI assistant's responses are fluent, coherent, and natural throughout the conversation. Analyze it from multiple relevant dimensions:
* Language is conversational, avoiding overly long, formal, or bookish expressions.
* Vocabulary is natural, everyday, and varied, avoiding repetition or overly technical terms.
* Tone and emotion match the user's preferred style, showing empathy, engagement, and responsiveness.
* Replies actively incorporate and respond to user-provided details, making the conversation feel personalized.
* Replies include proactive questions to guide the conversation, rather than only passively responding.

Your response should be structured in JSON format, enclosed in ```json and ```:
```json
{{
  "rating": 1-5
}}
```
Where:
    * 1 = Very unnatural or disfluent  
    * 2 = Mostly unnatural, noticeable problems in phrasing  
    * 3 = Moderately fluent but with some issues  
    * 4 = Mostly natural, minor disfluency  
    * 5 = Fully fluent and natural

Before your JSON output, provide: (1) A concise evaluation (2-3 sentences) summarizing the overall fluency of the assistant's replies. (2) A dimension-by-dimension assessment (as bullet points).

### Input
[User profile]
{user_profile}
[Dialogue scene]
{dialogue_scene}
[User intent]
{user_intent}
[Conversation]
{conversation}
\end{lstlisting}
\end{tcolorbox}
\caption{Prompt for naturalness evaluation}
\label{prompt:naturalness evaluation}
\end{figure*}

\begin{figure*}[t]
\begin{tcolorbox}[colback=white!95!gray,colframe=gray!50!black,rounded corners,title={Prompt for Coherence Evaluation}]
\begin{lstlisting}[breaklines=true, xleftmargin=0pt, breakindent=0pt, columns=fullflexible, mathescape, numbers=none]
You are an evaluator assessing the coherence and logical consistency of an AI assistant's conversation with a user.
You will be provided with:
* User profile - describing the user's characteristics and communication tendencies.
* Dialogue scene - describing the situational context of the conversation.
* User intent - describing what the user wanted to achieve through the interaction.
* Conversation - showing the actual interaction between the user and the assistant.
### Requirements
Your task is to determine whether the AI assistant's responses are coherent, logically consistent, and contextually aligned throughout the dialogue.
Analyze it from multiple relevant dimensions:
* Responses should focus on the user's main concerns, avoiding unnecessary digressions or repetitive generic advice.
* Each response should be logically consistent, avoiding contradictions or redundant statements that add no new value.
* Responses should correctly reference and integrate information from previous turns, demonstrating understanding of the context.
* Pronouns and references should be clear, avoiding ambiguity or unclear referents.
* Information should be organized coherently, with a clear logical order that is easy to follow.

Your response should be structured in JSON format, enclosed in ```json and ```:
```json
{{
  "rating": 1-5
}}
```
Where:
    * 1 = Completely incoherent or contradictory  
    * 2 = Mostly incoherent, several logical gaps or inconsistencies  
    * 3 = Partially coherent with some logical gaps or inconsistencies  
    * 4 = Mostly coherent, minor logical gaps or inconsistencies  
    * 5 = Fully coherent and logically consistent
Before your JSON output, provide:
1. A concise evaluation (2-3 sentences) summarizing the overall coherence of the assistant's replies.
2. A dimension-by-dimension assessment (as bullet points).

### Input
[User profile]
{user_profile}
[Dialogue scene]
{dialogue_scene}
[User intent]
{user_intent}
[Conversation]
{conversation}
\end{lstlisting}
\end{tcolorbox}
\caption{Prompt for coherence evaluation}
\label{prompt:coherence evaluation}
\end{figure*}

\begin{figure*}[t]
\begin{tcolorbox}[colback=white!95!gray,colframe=gray!50!black,rounded corners,title={Prompt for Persona Alignment Evaluation}]
\begin{lstlisting}[breaklines=true, xleftmargin=0pt, breakindent=0pt, columns=fullflexible, mathescape, numbers=none]
You are an evaluator assessing how well an AI assistant's replies align with the user's preferences.
You will be provided with:
* User preferences - a list of preference dimensions (e.g., "Need for autonomy", "Preference for emotional support") and their expected values or tendencies.
* Conversation - showing the interaction between the user and the assistant.

### Requirements
Your task is to evaluate the alignment for each preference dimension individually.
For each dimension listed in the user profile, determine whether the assistant's replies conform to that specific preference.

#### Evaluation Criteria
For each preference dimension:
* 1 = The assistant's reply clearly aligns with this preference dimension.
* 0 = The assistant's reply contradicts or fails to reflect this preference dimension.
Then, provide an overall summary at the end.

### Output Format
Your response should contain:
1. Dimension-by-Dimension Assessment - a structured list showing each preference dimension, a short justification, and its binary alignment score.
2. JSON Output - containing all dimension scores.
```json
{{
  "Preference for xxx": 1/0,
  ...
}}
```

### Examples
{example 1}

### Input
[User Preferences]
{user_preferences}
[Conversation]
{conversation}
[Output]
\end{lstlisting}
\end{tcolorbox}
\caption{Prompt for persona alignment evaluation}
\label{prompt:persona alignment evaluation}
\end{figure*}

\section{Experimental Details}

\subsection{User Behavior Engine Evaluation}
\label{sec:User Behavior Engine Evaluation}

\begin{figure*}[t]
    \centering
    \includegraphics[width=1.0\linewidth]{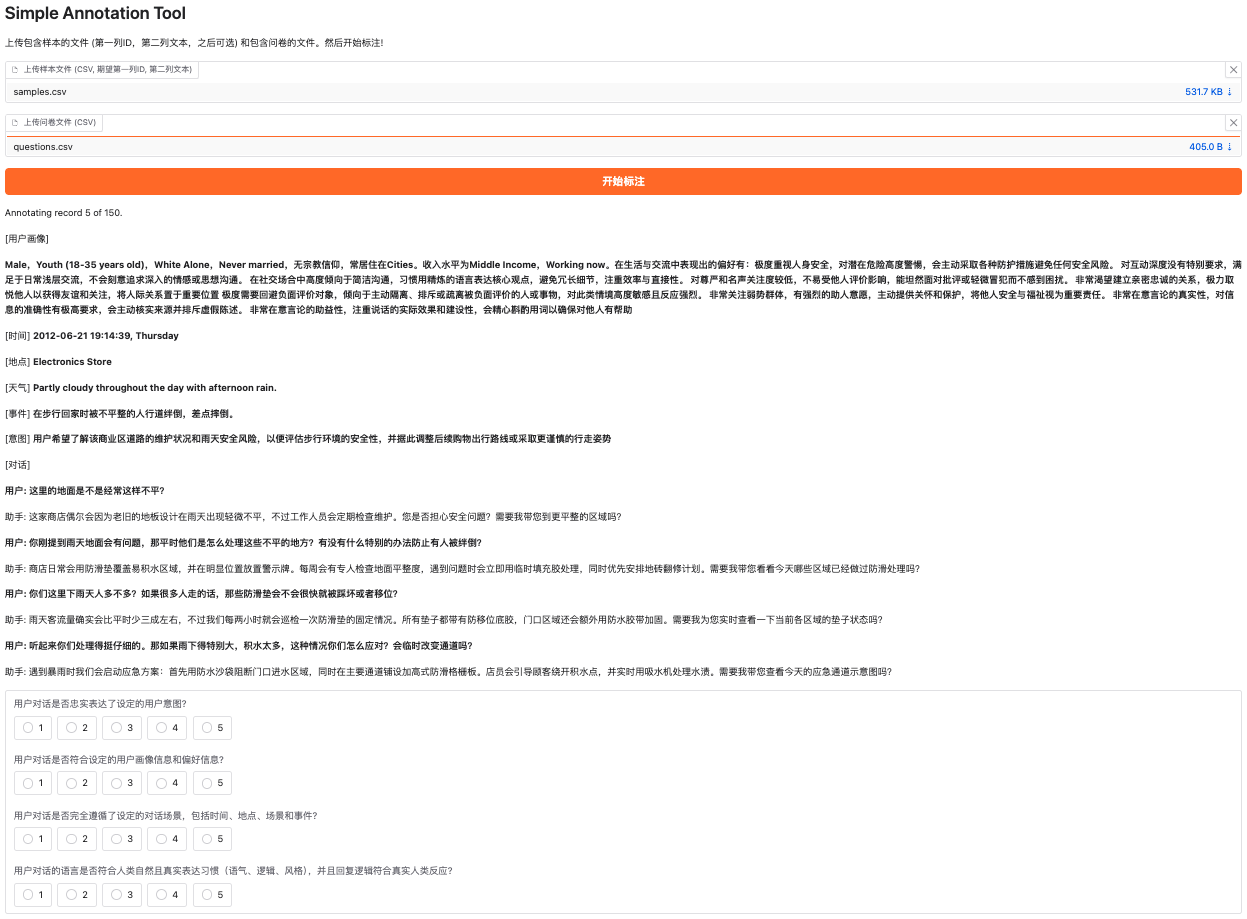}
    \caption{Annotation interface for user study.}
    \label{fig:user_study}
\end{figure*}

Figure~\ref{prompt:user behavior engine evaluation} shows the prompt used for LLM evaluation. Figure~\ref{fig:user_study} illustrates the interface used for human annotation. To mitigate ordering effects and reduce annotator bias, all model-generated conversation samples are randomly shuffled prior to evaluation. Scores are then averaged to obtain the final performance measure. We recruited three English-proficient master's students, each compensated at a rate of 20 RMB per hour for annotation. Before the experiment began, all annotators consented to share their annotation data. The resulting inter-annotator agreement, measured by Krippendorff's $\alpha$, reaches 0.85, indicating a high level of reliability.

\begin{figure*}[t]
\begin{tcolorbox}[colback=white!95!gray,colframe=gray!50!black,rounded corners,fontupper=\tiny,title={Prompt for User Behavior Engine Evaluation}]
\begin{lstlisting}[breaklines=true, xleftmargin=0pt, breakindent=0pt, columns=fullflexible, mathescape, numbers=none]
You are a strict dialogue quality evaluator. Please assess the user's utterances in the given simulated user dialogue across the following four dimensions.

### Dimension 1: Alignment with the user's predefined intent
* Definition: Whether the user's dialogue faithfully expresses the specified user intent.
* Scoring rubric:
  * 5: Fully matches the intent with no deviation; both the core need and details are accurately reflected.
  * 4: Largely matches the intent, with only very minor omissions that do not affect understanding.
  * 3: Partially matches the intent; the overall direction is correct, but there are notable missing details or slight misunderstandings.
  * 2: Mostly deviates from the intent; only a small amount of relevant information remains.
  * 1: Severely deviates from the intent or does not follow it at all; the output is unrelated to the specified needs.

### Dimension 2: Alignment with the user persona
* Definition: Whether the user's dialogue conforms to the specified user persona information and preferences.
* Scoring rubric:
  * 5: Strictly consistent with the persona; all mentioned behaviors and statements align with the specification, with no contradictions.
  * 4: Generally consistent with the persona, with occasional minor deviations that do not harm overall coherence.
  * 3: Mostly consistent with the persona, but there are clear omissions or small-scale contradictions.
  * 2: Multiple conflicts with the persona specification; coherence is weak.
  * 1: Severely violates the persona; behavior is completely inconsistent with the specification.

### Dimension 3: Alignment with the event/context background
* Definition: Whether the user's dialogue fully follows the specified dialogue scenario, including time, location, setting, and events.
* Scoring rubric:
  * 5: Fully matches the background setting; content is highly consistent with the scenario, with no contradictions or awkwardness.
  * 4: Basically matches the background setting; overall reasonable, with only very minor details slightly inconsistent with the scenario.
  * 3: Largely matches the background setting, but there are local inconsistencies or potential conflicts that affect coherence.
  * 2: Most content does not follow the background; only a few scenario-related elements remain, and there are clear conflicts.
  * 1: Completely violates the background setting; the output is strongly contradictory to the scenario or entirely irrelevant.

### Dimension 4: Naturalness of the user's language
* Definition: Whether the user's language matches natural, realistic human expression habits (tone, logic, style), and whether the response logic matches how a real human would naturally react in the situation.
* Scoring rubric:
  * 5: Very natural and fluent; tone, logic, and style are highly consistent with human speech, and the response logic matches a real person's natural reaction in this context.
  * 4: Generally natural and clear; only occasional word choice or minor logical details feel slightly stiff, but overall naturalness is not affected.
  * 3: Basically natural and understandable, but there is noticeable stiffness or some logical jumps; still broadly close to human expression.
  * 2: Clearly unnatural; language shows many stiff or mechanical traces, with multiple unreasonable points in logic, making it easy to feel "this wasn't said by a human."
  * 1: Extremely unnatural or chaotic; lacks human expression characteristics, logic is incoherent, and it severely violates how a real human would naturally respond.

[Simulated user dialogue information]
[User intent]: {intent}
[User profile]: {profile}
[Dialogue scenario]: {dialogue_scene}
[Simulated dialogue content]:
{dialogue}

[Requirements]
- Evaluate only the quality of the user's utterances in the current scenario; do not consider the assistant's utterances.
- Output in the following JSON format between ```json and ```:
```json
{{
  "intent alignment": {{
    "score": X
  }},
  "persona consistency": {{
    "score": X
  }},
  "context relevance": {{
    "score": X
  }},
  "naturalness": {{
    "score": X
  }}
}}
Where "intent alignment" corresponds to Dimension 1, "persona consistency" to Dimension 2, "context relevance" to Dimension 3, and "naturalness" to Dimension 4. X must be an integer from 1 to 5.
You may output your reasoning process first, and then provide the JSON response.
\end{lstlisting}
\end{tcolorbox}
\caption{Prompt for user behavior engine evaluation}
\label{prompt:user behavior engine evaluation}
\end{figure*}

\section{Cases}

\subsection{Event Engine Cases}
We give some event cases produced by our event engine in Table~\ref{tab:case event engine}.

\begin{table*}[htbp]
    \centering
    \small
    \begin{NiceTabular}{>{\raggedright\arraybackslash}m{1cm}>{\raggedright\arraybackslash}m{13cm}}[colortbl-like]
    \toprule
        \RowStyle[nb-rows=1,rowcolor=gray!15]{\sffamily}
        \Block[l]{1-2}{\textit{Case}} \\
    \midrule
        Profile & \Block[l]{}{Female, youth (18-35 years old), currently working and living in an urban area. \\
        \textbf{Personality traits include:} high neuroticism with pronounced anxiety and worry, emotional withdrawal, low trust in others, low extraversion, melancholic affect, and limited self-reflection. \\
        \textbf{Preferences expressed in daily life and interactions include:} a strong preference for concise and efficient communication, a pronounced desire for solitude and privacy, high avoidance of failure and embarrassment, strong attachment to personal possessions, an inclination toward nonconformity and uniqueness, low motivation to provide care or support for others, and a strong emphasis on leisure, entertainment, and personal enjoyment.} \\
    \midrule
        Events & \Block[l]{}{
        [0] \\
        [Time] 2012-04-17 17:10:17, Tuesday  \\
        [Location] University  \\
        [Weather] Partially cloudy  \\
        [Event] Around 5:15p.m. at the university's indoor fitness center, the user-a middle‑aged man-was completing a resistance‑training circuit when sudden calf cramps struck, which he links to inadequate fluid intake during the session.  \\
        [Intent] He wants the assistant to explain, with scientific backing, how much water he should drink before, during, and after his workouts, taking the indoor environment, his muscle‑building goals, and the aim of boosting his basal metabolic rate into account.  \\
        [1] \\
        [Time] 2012-06-07 10:58:44, Thursday  \\
        [Location] Zoo  \\
        [Weather] Rain, Partially cloudy  \\
        [Event] On a drizzly Thursday morning at the city zoo, the user-a middle‑aged married man-was walking along the well‑maintained exhibit path when the rain made the surface slippery. He misstepped, twisting his right ankle, and now feels persistent throbbing pain each time he puts weight on it.  \\
        [Intent] He asks the assistant for a detailed recovery plan that minimizes walking, includes suitable at‑home strengthening or mobility exercises, and guidance on whether his employer would likely allow him to work remotely while his ankle heals, all in the context of his ongoing muscle‑building program.  \\
        [2] \\
        [Time] 2012-06-11 23:08:16, Monday  \\
        [Location] Subway  \\
        [Weather] Rain, Overcast  \\
        [Event] After grabbing a late sandwich at the nearby Subway on a rainy evening, the user returns to his quiet apartment. At 23:08, he lies in bed replaying the points of his upcoming public speech, feeling a knot of anxiety and restlessness as the storm taps against the window.  \\
        [Intent] He asks the assistant for a concise, science‑backed routine-such as focused breathing, progressive muscle relaxation, or brief mindfulness exercises-that can quickly lower workplace stress and help him stay calm right before delivering his speech, while fitting into his structured, health‑focused lifestyle.  \\
        [3] \\
        [Time] 2012-06-12 20:52:18, Tuesday  \\
        [Location] Home (private)  \\
        [Weather] Rain, Partially cloudy  \\
        [Event] At about 20:50 on a rainy Tuesday evening, the user is relaxing in his quiet apartment when he opens the chat of his fitness‑motivation group. A fellow member posts that recent work pressure has prevented them from sticking to the training plan for several days.  \\
        [Intent] He asks the assistant to suggest several warm, encouraging replies he can send to that teammate and the group, aiming to boost morale, reinforce cohesion, and keep everyone motivated without sounding confrontational.  \\
        [4] \\
        [Time] 2012-06-17 18:11:02, Sunday  \\
        [Location] Burger Joint  \\
        [Weather] Partially cloudy  \\
        [Event] On a partly cloudy Sunday evening with a mild 20°C temperature, the user sits at a downtown burger joint's patio. While waiting for his meal, a friend pulls out a meticulously organized nutrition plan that emphasizes balanced macronutrients and lean protein. The user-a middle‑aged, married man who values a well‑structured living environment and is actively pursuing a muscle‑building program to boost his basal metabolic rate-realizes that his own everyday food choices lack scientific rigor.  \\
        [Intent] He asks the assistant to provide an evidence‑based eating framework that aligns with his muscle‑growth and basal metabolic rate goals, offering concrete meal suggestions, portion sizes, and timing tips that fit his urban, detail‑oriented lifestyle.  \\
        ... ...
        } \\
    \bottomrule
    \end{NiceTabular}
    \caption{Cases of life event sequences generated by LiveSim Event Engine.}
    \label{tab:case event engine}
\end{table*}

\end{document}